\documentclass[10pt,twocolumn,letterpaper]{article}
\usepackage[pagenumbers]{cvpr} % To force page numbers, e.g. for an arXiv version

\usepackage[dvipsnames]{xcolor}

\definecolor{cvprblue}{rgb}{0.21,0.49,0.74}
\usepackage[pagebackref,breaklinks,colorlinks,citecolor=cvprblue]{hyperref}

\usepackage{amssymb}
\usepackage{float}
\usepackage{adjustbox}
\usepackage{multicol}
\usepackage{multirow}
\usepackage{rotating}
\usepackage{algorithm}
\usepackage{algpseudocode}
\usepackage[titletoc]{appendix}

\newcommand{\cut}[1]{}

\title{DemoFusion: Democratising High-Resolution Image Generation With No \$\$\$}

\author{
Ruoyi Du$^{1,4\dagger}$, Dongliang Chang$^{2*}$, Timothy Hospedales$^3$, Yi-Zhe Song$^4$, Zhanyu Ma$^1$\\
$^1$PRIS, Beijing University of Posts and Telecommunications, China\\ $^2$Tsinghua University, China \quad $^3$University of Edinburgh, UK \quad $^4$SketchX, University of Surrey, UK\\
{\tt\small \{duruoyi, mazhanyu\}@bupt.edu.cn},
{\tt\small changdongliang@pris-cv.cn}, \\
{\tt\small t.hospedales@ed.ac.uk}, 
{\tt\small y.song@surrey.ac.uk} \\
{\small \url{https://ruoyidu.github.io/demofusion/demofusion.html}}
\vspace{-0.7cm}
}

\begin{document}

% \twocolumn[{
% \maketitle
% \vspace{-0.5cm}
% \begin{figure}[H]
% \hsize=\textwidth
% \centering
% \includegraphics[width=2\linewidth]{figures/illustration.pdf}
% \captionof{figure}{}
% \vspace{-0.2cm}
% \end{figure}
% }]
\twocolumn[{
    \renewcommand\twocolumn[1][]{#1}
    \maketitle
    \begin{center}
        \centering
        % \vspace{-0.6cm}
        \includegraphics[width=\textwidth]{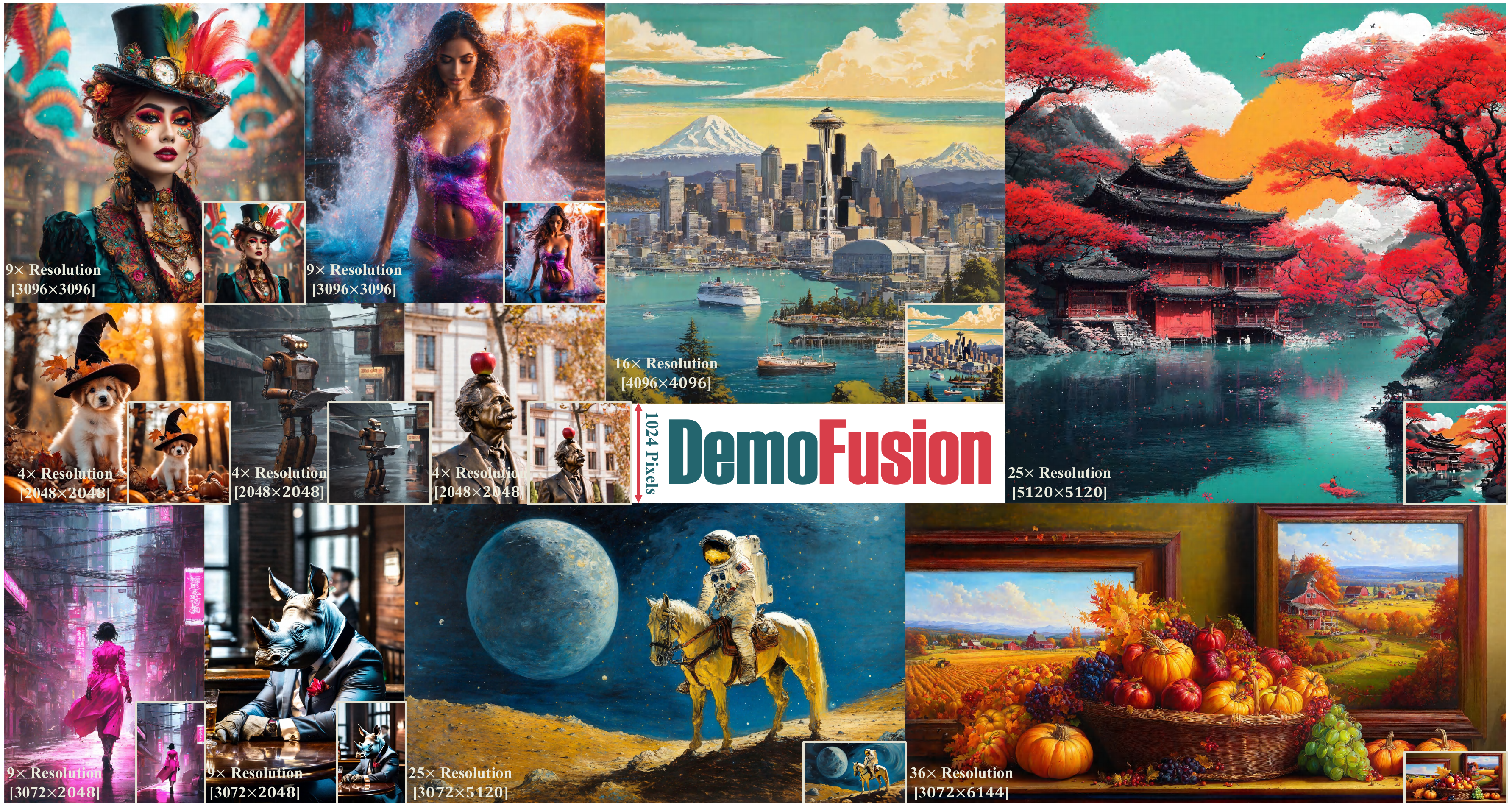}
        \captionof{figure}{\textbf{Selected landscape samples of DemoFusion \emph{versus} SDXL}~\cite{podell2023sdxl} (all images in the figure are presented at their actual sizes). SDXL can synthesize images up to a resolution of $1024^2$, while DemoFusion extends SDXL to generate images at $4\times$, $16\times$, and even higher resolutions without any fine-tuning or prohibitive memory demands. All generated images are produced using a single RTX $3090$ GPU. Best viewed \textbf{ZOOMED-IN}.}
        \label{fig:illustration}
        % \vspace{-0.cm}
    \end{center}
}]

\footnotetext{$^\dagger$The work is done while Ruoyi Du visiting the People-Centred AI Institute at the University of Surrey}
\footnotetext{$^*$Corresponding Author}
% \maketitle
\begin{abstract}
\vspace{-0.3cm}
High-resolution image generation with Generative Artificial Intelligence (GenAI) has immense potential but, due to the enormous capital investment required for training, it is increasingly centralised to a few large corporations, and hidden behind paywalls. This paper aims to democratise high-resolution GenAI by advancing the frontier of high-resolution generation while remaining accessible to a broad audience. We demonstrate that existing Latent Diffusion Models (LDMs) possess untapped potential for higher-resolution image generation. Our novel DemoFusion framework seamlessly extends open-source GenAI models, employing Progressive Upscaling, Skip Residual, and Dilated Sampling mechanisms to achieve higher-resolution image generation. The progressive nature of DemoFusion requires more passes, but the intermediate results can serve as ``previews'', facilitating rapid prompt iteration.
\vspace{-0.4cm}

\end{abstract}    
\section{Introduction}
Generating high-resolution images with Generative Artificial Intelligence (GenAI) models has demonstrated remarkable potential~\cite{OpenAI2021DALL-E,MidJourney2022,StableDiffusion2022}. However, these capabilities are increasingly centralised. Training high-resolution image generation models requires substantial capital investments in hardware, data, and energy that are beyond the reach of individual enthusiasts and academic institutions. For example, training Stable Diffusion $1.5$, at a resolution of $512^2$, entails over $20$ days of training on $256$ A$100$ GPUs\footnote{Information from: \url{https://huggingface.co/runwayml/stable-diffusion-v1-5}}. Companies that make these investments understandably want to recoup their costs and increasingly hide the resulting models behind paywalls. This trend toward centralisation and pay-per-use access is accelerating as GenAI image synthesis advances in quality since the investment required to train image generators increases rapidly with image resolution.

In this paper we reverse this trend and re-democratise GenAI image synthesis by introducing \emph{DemoFusion}, which pushes the frontier of high-resolution image synthesis from $1024^2$ in  SDXL~\cite{podell2023sdxl}, Midjourney~\cite{MidJourney2022}, DALL-E~\cite{OpenAI2021DALL-E}, \emph{etc} to $4096^2$ or more. DemoFusion requires no additional training and runs on a single consumer-grade RTX $3090$ GPU (hardware for the ``working class'' in the GenAI era), as shown in Fig.~\ref{fig:illustration}. The only trade-off? A little more patience.

\begin{figure}[t]
\centering
\includegraphics[width=1\linewidth]{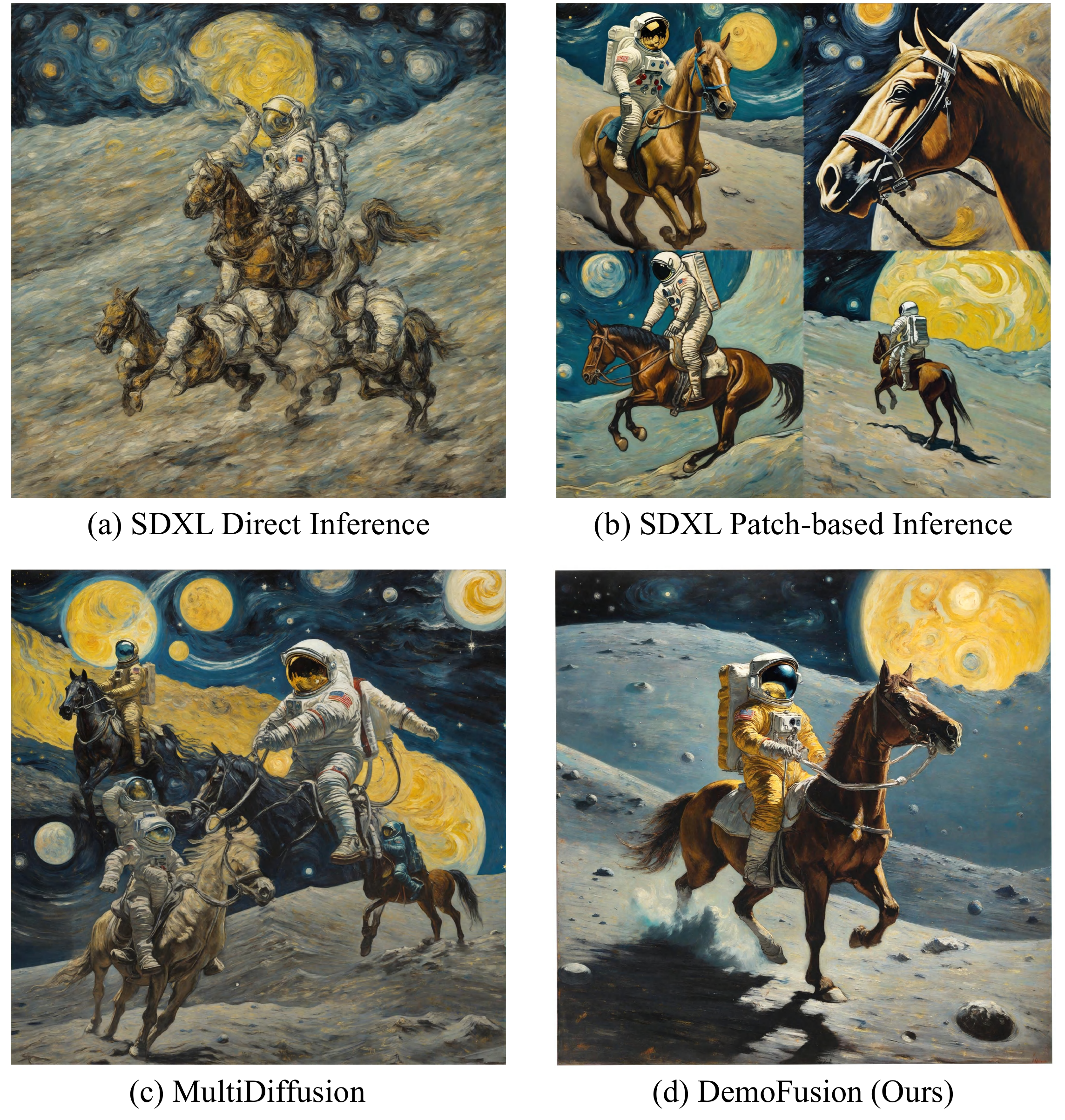}
\caption{\textbf{Examples of $4\times$ ($2048^2$) generation based on SDXL}~\cite{podell2023sdxl}. (a) Directly prompting SDXL to generate a $4\times$ image. (b) SDXL~\cite{podell2023sdxl}  inferences on non-overlapping patches at the original resolution. It fails, but reveals that the SDXL possesses prior knowledge of localized patches at higher resolutions. (c) MultiDiffusion~\cite{bar2023multidiffusion} fuses multiple overlapping denoising paths to generate higher-resolution images without edge effects, but lacks the global context for semantic coherence. (d) Our proposed DemoFusion achieves global semantic coherence in high-resolution generation.}% through the techniques of Progressive Upscaling, Skip Residual, and Dilated Sampling, enbales local denoising paths to serve the global context.}}
\label{fig:intuition}
\vspace{-0.5cm}
\end{figure}

Specifically, we start with the open source SDXL~\cite{podell2023sdxl} model, capable of generating images of $1024^2$. DemoFusion is a plug-and-play extension to SDXL that enables $4\times$, $16\times$, or more increase in generation resolution (Fig~\ref{fig:illustration}) -- all with zero additional training, and only a few simple lines of code. Off-the-shelf SDXL fails if directly prompted to produce higher-resolution images (Fig.~\ref{fig:intuition} (a)). However, we observe that text-to-image LDMs encounter many cropped photos during their training process. These cropped photos either exist inherently in the training set or are intentionally cropped for data augmentation. Consequently, models like SDXL occasionally produce outputs that focus on localised portions of objects~\cite{podell2023sdxl}, as illustrated in Fig.~\ref{fig:intuition} (b). In other words, existing open-source LDMs already contain sufficient prior knowledge to generate high-resolution images, if only we can unlock them by fusing multiple such high-resolution patches into a complete scene. 

However, achieving coherent patch-wise high-resolution generation is non-trivial. A recent study, MultiDiffusion~\cite{bar2023multidiffusion} showcased the potential of fusing multiple overlapped denoising paths to generate panoramic images. Yet, when directly applying this approach to generate specific high-resolution object-centric images, results are repetitive and distorted without global semantic coherence \cite{zheng2023any}, as illustrated in Fig.~\ref{fig:intuition} (c). We conjecture the underlying reason is that overlapped patch denoising merely reduces the seam issue without a broad perception of the global context required for semantic coherence.  
DemoFusion builds upon the same idea of fusing multiple denoising paths from a pre-trained SDXL model to achieve high-resolution generation. It introduces three key mechanisms to achieve global semantic coherence together with rich local detail (Fig.~\ref{fig:intuition} (d) vs (a, c)): (i) \textit{Progressive Upscaling}: Starting with the low-resolution input, DemoFusion iteratively enhances images through an ``upsample-diffuse-denoise" loop, using the noise-inversed lower-resolution image as a better initialisation for generating the higher-resolution image. (ii) \textit{Skip Residual}: Within the same iteration, we additionally utilise the intermediate noise-inversed representations as skip residuals, maintaining global consistency between high and low-resolution images. (iii) \textit{Dilated Sampling}: We extend MultiDiffusion to increase global semantic coherence by using dilated sampling of denoising paths. These three techniques to modify inference are simple to implement on a pre-trained SDXL and provide a dramatic boost in high-resolution image generation quality and coherence. Fig.~\ref{fig:framework} illustrates the framework.

The caveat is that generating high-resolution images does require more runtime (users need to exercise more patience). This is partially due to the progressive upscaling requiring more passes; however, primarily because the time required grows exponentially with resolution (as per any patch-wise LDM \cite{bar2023multidiffusion}), and thus, the highest resolution pass dominates the cost. Nevertheless, the memory cost is low enough for consumer-grade GPUs, and progressive generation allows the users to preview low-resolution results rapidly, facilitating rapid iteration on the prompt until satisfaction with the general layout and style, prior to waiting for a full high-resolution generation. 

%However, because the time required for generation grows exponentially with resolution, the time taken by the passes at lower resolutions is not as significant compared to the unavoidable time required for the target resolution. Moreover, progressive generation .}

\section{Related Work}
With the progress of several years, diffusion model (DM)~\cite{sohl2015deep} has recently reached its own ``tipping point'' -- with the emergence of works like DDPM~\cite{song2020denoising}, DDIM~\cite{song2020denoising}, ADM~\cite{dhariwal2021diffusion}, DM has shown great potential in image generation due to its outstanding generation quality and diversity. Subsequently, using a pre-trained autoencoder, the latent diffusion model (LDM)~\cite{rombach2022high} applies a diffusion model in the latent space, achieving efficient training and inference. This enabled the emergence of high-performance generative models trained on billions of data, such as the Stable Diffusion series. LDM's excellent generalisation capability has led to subsequent research on controllable generation~\cite{zhang2023adding,mou2023t2i,ruiz2023dreambooth} and editable generation~\cite{hertz2022prompt,mokady2023null,brooks2023instructpix2pix}; it has also been widely applied in numerous downstream generative tasks, such as text-to-video~\cite{ho2022imagen,wu2023tune,he2023latent}, text-to-$3$D~\cite{poole2022dreamfusion,lin2023magic3d,xu2023dream3d}, text-to-avatar~\cite{wang2023rodin,han2023headsculpt,liao2023tada}, and text-to-human sketch~\cite{jain2023vectorfusion,qu2023sketchdreamer}, \emph{etc}.

Despite achieving numerous successes, current LDMs like Stable Diffusion $1.5$ and Stable Diffusion XL are still confined to generating images at resolutions of $512^2$ and $1024^2$, respectively~\cite{podell2023sdxl}. Escalating resolution significantly increases training expenses and computational load, making such models impractical for most researchers and users. An intuitive solution to generate high-resolution images involves using LDMs for initial image generation, followed by enhancement through a super-resolution (SR) model. Cascaded Diffusion Models~\cite{ho2022cascaded} cascades several diffusion-based SR models behind a diffusion model, but its application remains capped at $256^2$ resolution images. We attempted to enhance state-of-the-art LDMs with SR models~\cite{zhang2021designing,wang2023exploiting}, but found that images generated at lower resolutions were deficient in detail. Upscaling these images with SR failed to yield the high-resolution detail desired. Another attempt is to retrain/fine-tune open-source DMs to achieve satisfactory results~\cite{hoogeboom2023simple,zheng2023any}, but fine-tuning still brings a non-negligible cost.

Recently, MultiDiffusion~\cite{bar2023multidiffusion} fuses multiple overlapped denoising paths of LDMs, achieving seamless panorama generation in a training-free manner. Subsequently, SyncDiffusion~\cite{lee2023syncdiffusion} further constrains the consistency between denoising paths using a gradient descent approach. However, these methods are limited to generating scene images through repetition; when applied to generating specific objects, they lead to local repetition and structural distortion. Valuing the training-free characteristic of such methods, we proposed DemoFusion based on MultiDiffusion in this paper towards democratising high-resolution generation.

Note that a recent concurrent work, SCALECRAFTER~\cite{he2023scalecrafter}, with the same motivation, proposed a tuning-free framework for high-resolution image generation. It ingeniously adapts the diffusion model for higher resolutions by dilating its convolution kernels at specific layers. Despite a smart move, our experiments indicate that SCALECRAFTER somewhat degrades the model's performance and does not bring about the local details expected at higher resolutions. In contrast, DemoFusion has demonstrated better results.
\section{Methodology}
\begin{figure*}[h]
\centering
\includegraphics[width=1\linewidth]{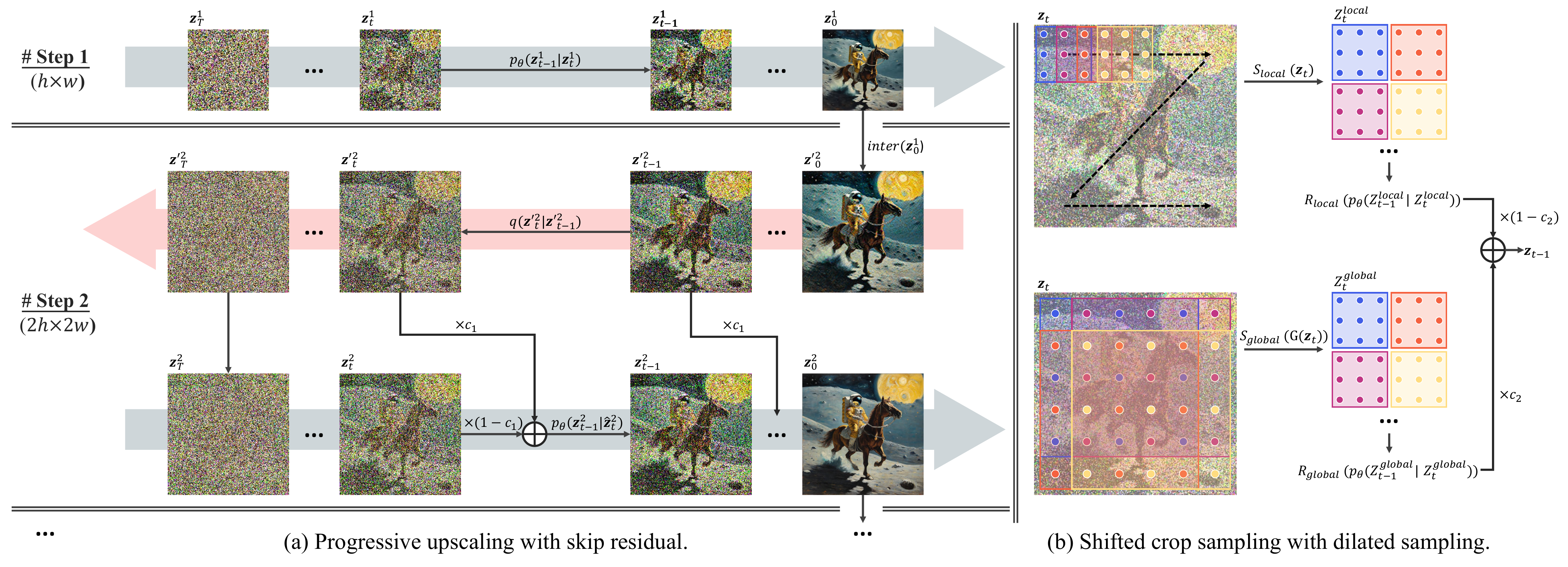}
\vspace{-0.6cm}
\caption{\textbf{The proposed DemoFusion framework.} (a) Starting with conventional resolution generation, DemoFusion engages an ``upsample-diffuse-denoise'' loop, taking the low-resolution generated results as the initialization for the higher resolution through noise inversion. Within the ``upsample-diffuse-denoise'' loop, a noise-inverted representation from the corresponding time-step in the preceding diffusion process serves as skip-residual as global guidance. (b) To improve the local denoising paths of MultiDiffusion, we introduce dilated sampling to establish global denoising paths, promoting more globally coherent content generation.}
\label{fig:framework}
\vspace{-0.4cm}
\end{figure*}

\subsection{Preliminaries}
\noindent\textbf{Latent Diffusion Model}: Given an image $\mathbf{x}$, an LDM first encodes it to the latent space via the encoder of the pre-trained autoencoder, \emph{i.e.}, $\mathbf{z} = \mathcal{E}(\mathbf{x}), \mathbf{z} \in \mathbb{R}^{c \times h \times w}$.

Following this, the two core components of the diffusion model, the diffusion and the denoising process, take place in the latent space. The diffusion process comprises a sequence of $T$ steps with Gaussian noise incrementally introducing into the latent distribution at each step $t \in [0, T]$. With a prescribed variance schedule $\beta_1, \cdots, \beta_T$, the diffusion process can be formulated as
\begin{equation}~\label{}
    q(\mathbf{z}_t|\mathbf{z}_{t-1})=\mathcal{N}(\mathbf{z}_t;\sqrt{1-\beta_t} \mathbf{z}_{t-1}, \beta_t \mathbf{I}).
\end{equation}
In contrast, the denoising process aims to recover the cleaner version $\mathbf{z}_{t-1}$ from $\mathbf{z}_t$ by estimating the noise, which can be expressed as 
\begin{equation}~\label{}
    p_\theta(\mathbf{z}_{t-i}|\mathbf{z}_t)=\mathcal{N}(\mathbf{z}_{t-1};\mathbf{\mu}_\theta(\mathbf{z}_t, t), \mathbf{\Sigma}_\theta(\mathbf{z}_t, t)),
\end{equation}
where $\mathbf{\mu}_\theta$ and $\mathbf{\Sigma}_\theta$ are determined through estimation procedures and $\theta$ denotes the parameters of the denoise model.

\noindent\textbf{MultiDiffusion}: MultiDiffusion \cite{bar2023multidiffusion} extends LDMs such as SDXL to produce high-resolution panoramas by overlapped patch-based denoising. 
% \cut{Here, we focus solely on one of its applications, panoramic image generation, and based on this, we expand to achieve high-resolution image generation.}

In simple terms, MultiDiffusion defines a latent space $\mathbb{R}^{c \times H \times W}$ with $H > h$ and $W > w$. For arbitrary denoising step $t$ with $\mathbf{z}_t \in \mathbb{R}^{c \times H \times W}$, MultiDiffusion first applies a shifted crop sampling $\mathcal{S}_{local}(\cdot)$ to obtain a series of local latent representations, \emph{i.e.}, $Z_t^{local} = [\mathbf{z}_{0,t}, \cdots, \mathbf{z}_{n,t}, \cdots, \mathbf{z}_{N,t}] = \mathcal{S}_{local}(\mathbf{z}_t)$, $\mathbf{z}_{n,t} \in \mathbb{R}^{c \times h \times w}$, where $N = (\frac{(H - h)}{d_h} + 1) \times (\frac{(W - w)}{d_w} + 1)$, $d_h$ and $d_w$ is the vertical and horizontal stride, respectively.

After that, the conventional denoising process is independently applied to these local latent representations via $p_\theta(\mathbf{z}_{n,t-1}|\mathbf{z}_{n,t})$. And then $Z_{t-1}^{local}$ is reconstructed to the original size with the overlapped parts averaged as $\mathbf{z}_{t-1}=\mathcal{R}_{local}(Z_{t-1}^{local})$, where $\mathcal{R}_{local}$ denotes the reconstruction process. Eventually, a higher-resolution panoramic image can be obtained by directly decoding $\mathbf{z}_0$ into image $\hat{\mathbf{x}}$. 

MultiDiffusion provides effective panorama generation, thanks to smoothing the edge effects between generated patches. However, as discussed by \cite{zheng2023any}, and illustrated in Fig.~\ref{fig:intuition}, it struggles with generating coherent semantic content for specific objects. The fundamental reason for this is that each patch/diffusion path is constrained only by the text condition and lacks awareness of the global context of the other patches. 

We introduce three modifications to the inference procedure of SDXL that enable a patch-wise high-resolution image generation strategy to achieve both global semantic coherence and rich local details. These are: \emph{Progressive Upscaling} (see Sec.~\ref{sec:progressive_generation}), \emph{Skip Residual} (see Sec.~\ref{sec:skip_residual}) and \emph{Dilated Sampling} (see Sec.~\ref{sec:dilated_sampling}). The overall flow of DemoFusion is summarised in Appendix~\ref{sec:pseudo_code}.

\subsection{Progressive Upscaling}~\label{sec:progressive_generation}
Progressively generating images from low to high resolution is a well-established concept~\cite{karras2018progressive}. By initially synthesizing a semantically coherent overall structure at low resolution, and subsequently increasing resolution to add detailed local features, models can produce coherent yet rich images. In this paper, we present a novel \emph{progressive upscaling} generation process tailored for LDMs (Fig~\ref{fig:framework} (a)).

Consider a pre-trained latent diffusion model with parameters $\theta$, operating on the latent space $\mathbb{R}^{c \times h \times w}$ to produce images with a resolution magnified by a factor of $K$. The scaling factor for the side length should be $S=\sqrt{K}$. And the target latent space is $\mathbb{R}^{c \times H \times W}$ where $H=Sh$ and $W=Sw$. {Instead of directly synthesizing $\mathbf{z}_t \in \mathbb{R}^{c \times H \times W}$, we break the generation process into $S$ distinct phases, each consisting of an ``upsample-diffuse-denoise'' loop, except for the first phase which follows an ``initialise-denoise'' scheme.} Specifically, given diffusion and denoising process as $q(\mathbf{z}_T|\mathbf{z}_{0}) = \prod_{t=1}^T q(\mathbf{z}_t|\mathbf{z}_{t-1})$ and $p_{\theta}(\mathbf{z}_0|\mathbf{z}_{T}) = \prod_{t=T}^1 p_{\theta}(\mathbf{z}_{t-1}|\mathbf{z}_t)$. Then, we can formulate the proposed progressive upscaling generation process as
\vspace{-0.2cm}
\begin{equation}
\begin{aligned}~\label{}
    p_\theta(\mathbf{z}^S_0|\mathbf{z}^1_T)&=p_{\theta}(\mathbf{z}^1_0|\mathbf{z}^1_{T}) \prod_{s=2}^S (q(\mathbf{z'}^s_T|\mathbf{z'}^s_{0}) p_{\theta}(\mathbf{z}^s_0|\mathbf{z'}^s_{T})),
\end{aligned}
\end{equation}
where $\mathbf{z'}^{s}_0$ is obtained through explicit upsampling as $\mathbf{z'}^{s}_0=inter(\mathbf{z}^{s-1}_0)$ and $inter(\cdot)$ is an arbitrary interpolation algorithm (\emph{e.g.}, bicubic). In essence, we first run a regular LDM such as SDXL as $p_{\theta}(\mathbf{z}^1_0|\mathbf{z}^1_{T})$. We then iteratively for each scale $s$: (i) upscale the low-resolution image $\mathbf{z}^{s-1}_0$ to $\mathbf{z'}^s_0$, (ii) reintroduce noise via the diffusion process to obtain $\mathbf{z'}^s_{T}$, and (iii) denoise to obtain $\mathbf{z}^s_0$. By repeating this process, we can compensate for the artificial interpolation-based upsampling and gradually fill in more and more local details.

%generate a low-resolution latent representation $\mathbf{z}^{s-1}_0$ through the denoising process, then upscale it to $\mathbf{z'}^s_0$, and reintroduce noise via the diffusion process to obtain $\mathbf{z'}^s_{T}$ as initialization for the next phase, \emph{i.e.}, $\mathbf{z}^s_{T}=\mathbf{z'}^s_{T}$. {By repeating this process, we can dilute the artificial noises introduced by interpolation through diffusing and reconstruct local details at higher resolutions through denoising.}

\subsection{Skip Residual}~\label{sec:skip_residual}
{The ``diffuse-denoise'' process has parallels in some image editing works -- people attempt to find the initial noise of an image using specialized noise inversion techniques, ensuring that the unedited parts remain consistent with the original image during the denoising editing process~\cite{hertz2022prompt,mokady2023null}. However, these inversion techniques are less practical to DemoFusion's denoising process. Therefore, we instead simply use a conventional diffusion process by adding random Gaussian noise.}

{However, directly diffusing $\textbf{z}^s_0$ to $\mathbf{z'}^s_T$ as initialization would result in most information loss. In contrast, diffusing to an intermediate $t$ and then starting denoise from $\mathbf{z'}^s_t$ might be better. However, it is challenging to determine the optimal intersection time-step $t$ of the ``upsample-diffuse-denoise'' loop -- the larger the $t$, the more information is lost, which weakens the global perception; the smaller the $t$, the stronger the noise introduced by upsampling (refer to Appendix~\ref{sec:discussion}). It is a difficult trade-off and could be example-specific. Therefore, we introduce the skip residual as a general solution, which can be informally considered as a weighted fusion of multiple ``upsample-diffuse-denoise'' loops with a series of different intersection time-steps $t$ (Fig.~\ref{fig:framework} (a)).}

For each generation phase $s$, we have already obtained a series of noise-inversed versions of $\mathbf{z'}^s_0$ as $\mathbf{z'}^s_t$ with $t \in [1, T]$. During the denoising process, we introduce the corresponding noise-inversed versions as \emph{skip residuals}. In other words, we modify $p_{\theta}(\mathbf{z}_{t-1}|\mathbf{z}_t)$ to $p_{\theta}(\mathbf{z}_{t-1}|\mathbf{\hat{z}}_t)$ with
\begin{equation}~\label{}
    \mathbf{\hat{z}}^s_t = c_1 \times \mathbf{z'}^s_t + (1-c_1) \times \mathbf{z}^s_t,
\end{equation}
where $c_1 = ((1 + \cos\left(\frac{T-t}{T}\times \pi \right)) / 2)^{\alpha_1}$ is a scaled cosine decay factor with a scaling factor $\alpha_1$. This essentially utilizes the results from the previous phase to guide the generated image's global structure during the initial steps of the denoising process. Meanwhile, we gradually reduce the impact of the noise residual, allowing the local denoising paths to optimize the finer details more effectively in the later steps.

\begin{figure*}[t]
\centering
\includegraphics[width=0.96\linewidth]{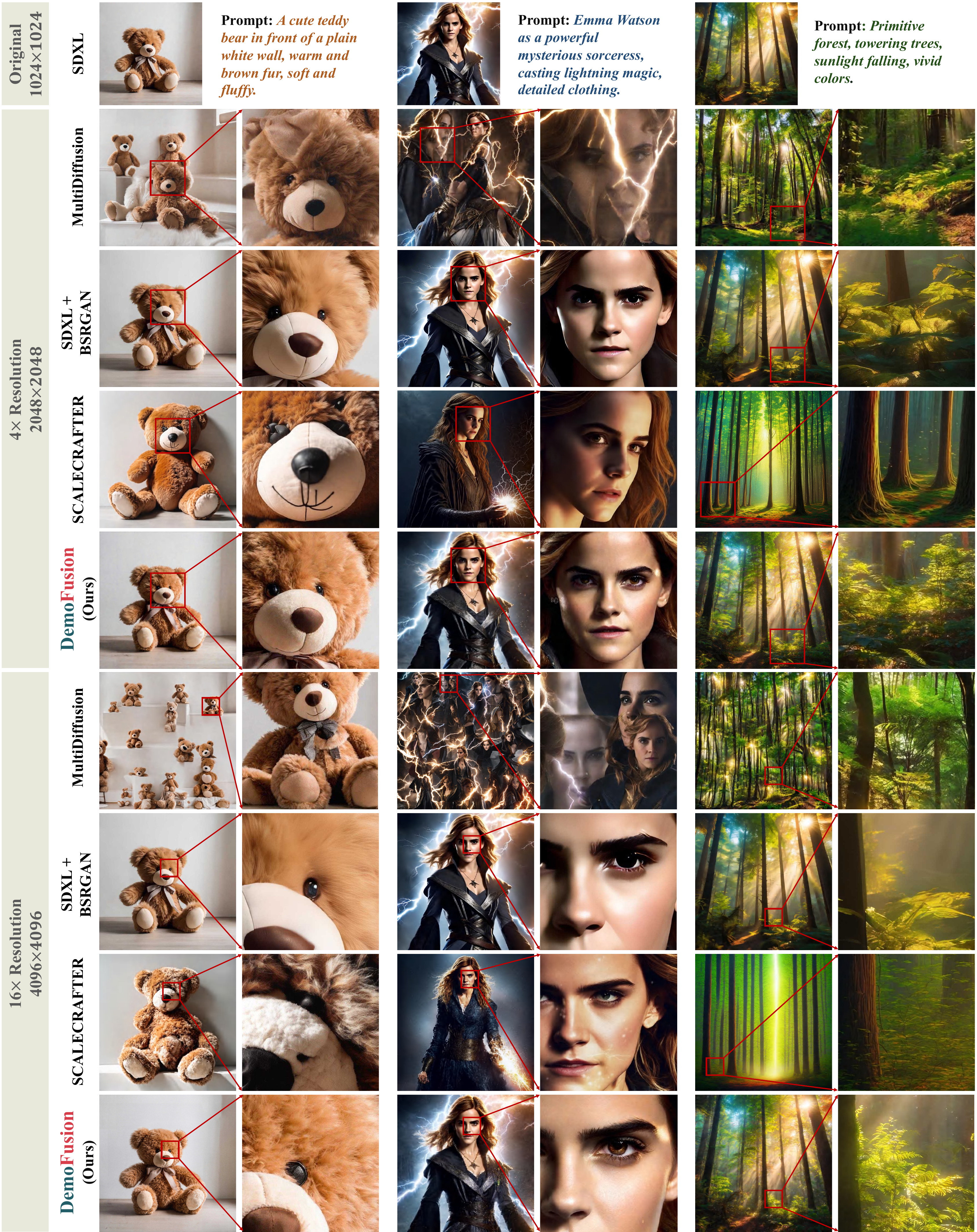}
\caption{\textbf{Qualitative comparison with other baselines.} Local details have already been zoomed in, but it's still recommended to \textbf{ZOOM IN} for a closer look.}
\label{fig:comparison}
\end{figure*}

\begin{table*}[t]
% 	\vspace{-1.0cm} 
\renewcommand{\arraystretch}{1.3}
\centering
\begin{adjustbox}{width=1\linewidth,center}
\begin{tabular}{c|ccccc|c|ccccc|c|ccccc|c}
\toprule[1pt]
\multirow{2}{*}{\textbf{Method}} & \multicolumn{6}{c|}{$2048 \times 2048$} & \multicolumn{6}{c|}{$2048 \times 4096$} & \multicolumn{6}{c}{$4096 \times 4096$}
\cr\cline{2-19} & FID $\downarrow$ & IS $\uparrow$ & FID$_{crop}$ $\downarrow$ & IS$_{crop}$ $\uparrow$ & CLIP $\uparrow$ & Time & FID $\downarrow$ & IS $\uparrow$ & FID$_{crop}$ $\downarrow$ & IS$_{crop}$ $\uparrow$ & CLIP $\uparrow$ & Time & FID $\downarrow$ & IS $\uparrow$ & FID$_{crop}$ $\downarrow$ & IS$_{crop}$ $\uparrow$ & CLIP $\uparrow$ & Time\\
% \midrule[0.5pt]
% \multicolumn{19}{c}{\emph{\textbf{Greedy-Hungarian}}}\\
\midrule[1pt]
SDXL Direct Inference~\cite{podell2023sdxl} & $79.66$ & $13.47$ & $73.91$ & $17.38$ & $28.12$ & $1$ min & $97.08$ & $14.12$ & $96.41$ & $18.01$ & $27.29$ & $3$ min & $105.65$ & $14.01$ & $98.59$ & $19.47$ & $25.64$ & $8$ min \\
MultiDiffusion~\cite{bar2023multidiffusion} & $75.93$ & $14.56$ & $70.93$ & $17.85$ & $28.97$ & $3$ min & $89.38$ & $14.17$ & $82.78$ & $18.87$ & $28.66$ & $6$ min & $97.98$ & $13.84$ & $79.45$ & $19.73$ & $28.62$ & $15$ min \\
SDXL $+$ BSRGAN~\cite{zhang2021designing} & $\underline{66.41}$ & $\underline{16.22}$ & $\underline{67.42}$ & $\underline{21.11}$ & $\underline{29.61}$ & $1$ min & $\mathbf{68.70}$ & $\underline{16.29}$ & $\underline{75.03}$ & $\underline{21.76}$ & $\underline{29.01}$ & $1$ min & $\mathbf{66.44}$ & $\mathbf{16.21}$ & $\underline{77.20}$ & $\underline{22.42}$ & $\mathbf{29.63}$ & $1$ min \\
SCALECRAFTER~\cite{he2023scalecrafter} & $69.91$ & $15.72$ & $68.36$ & $19.44$ & $29.51$ & $1$ min & $80.16$ & $15.29$ & $83.08$ & $19.56$ & $28.87$ & $6$ min & $87.50$ & $15.20$ & $84.36$ & $20.32$ & $29.04$ & $19$ min \\
DemoFusion (Ours) & $\mathbf{65.73}$ & $\mathbf{16.41}$ & $\mathbf{64.81}$ & $\mathbf{21.40}$ & $\mathbf{29.68}$ & $3$ min & $\underline{73.15}$ & $\mathbf{16.37}$ & $\mathbf{71.35}$ & $\mathbf{23.55}$ & $\mathbf{29.05}$ & $11$ min & $\underline{74.11}$ & $\underline{16.11}$ & $\mathbf{70.34}$ & $\mathbf{24.28}$ & $\underline{29.57}$ & $25$ min \\
\midrule[1pt]
\end{tabular}
\end{adjustbox}
\vspace{-0.2cm}
\caption{\textbf{Quantitative comparison results}. The best results are marked in \textbf{bold}, and the second best results are marked by \underline{underline}.}
\label{tbl:comparison}
\vspace{-0.1cm}
\end{table*}

\subsection{Dilated Sampling}~\label{sec:dilated_sampling}
Beyond the explicit integration of global information as a residual, we introduce \emph{dilated sampling} to give each denoising path more global context. The technique of dilating convolutional kernels to expand their receptive field is conventional in various dense prediction tasks~\cite{yu2018multi}. The concurrent tuning-free method, SCALECRAFTER~\cite{he2023scalecrafter}, similarly uses dilated convolutional kernels for adapting trained latent diffusion models to higher-resolution image generation. However, our approach diverges here: rather than dilating the convolutional kernel, we directly dilate the sampling within the latent representation. After that, the global denoising paths, derived through dilated sampling, are processed analogously to local denoising paths in MultiDiffusion.

As depicted in Fig.~\ref{fig:framework} (b), we applied shifted dilated sampling to obtain a series of global latent representation, \emph{i.e.}, $Z_t^{global} = [\mathbf{z}_{0,t}, \cdots, \mathbf{z}_{m,t}, \cdots, \mathbf{z}_{M,t}] = \mathcal{S}_{global}(\mathbf{z}_t)$, $\mathbf{z}_{m,t} \in \mathbb{R}^{c \times h \times w}$. To sample from the whole latent representation, the dilation factor is set to be $s$ and $M = s^2$. Similarly, we apply the general denosing process on these global latent representations as $p_\theta(\mathbf{z}_{m,t-1}|\mathbf{z}_{m,t})$. Then, the reconstructed global representations are fused with the reconstructed local representations to form the final latent representation:
\begin{equation}~\label{}
\mathbf{z}_{t-1}=c_2 \times \mathcal{R}_{global}(Z_{t-1}^{global}) + (1-c_2) \times  \mathcal{R}_{local}(Z_{t-1}^{local}),
\end{equation}
where $c_2 = ((1 + \cos\left(\frac{T-t}{T}\times \pi\right)) / 2)^{\alpha_2}$ is a scaled cosine decay factor with a scaling factor $\alpha_2$, also chosen based on the characteristic of the diffusion model where earlier steps mainly reconstruct the overall structure, while later steps focus on refining the details.

It is noteworthy that directly using dilated sampling can lead to grainy images. This is because, unlike the local denoising paths, which have overlaps, the global denoising paths operate independently of each other. To address this issue, we employ a straightforward yet intuitive approach -- applying a Gaussian filter $\mathcal{G}(\cdot)$ to the latent representation before performing dilated sampling as $Z_t^{global} = \mathcal{S}_{global}(\mathcal{G}(\mathbf{z}_t))$. The kernel size of the Gaussian filter is set to be $4s-3$, making it sufficient at every phase. Moreover, the standard deviation of the Gaussian filter will decrease from $\sigma_1$ to $\sigma_2$ as $c_3 \times (\sigma_1 - \sigma_2) + \sigma_2$, where $c_3 = ((1 + \cos\left(\frac{T-t}{T}\times \pi\right)) / 2)^{\alpha_3}$ is also a scaled cosine decay factor with a scaling factor $\alpha_3$, ensuring that the effect of the filter gradually diminishes as the directions of global denoising paths become consistent, preventing the final image from becoming blurry. 
\begin{figure*}[t]
\centering
\includegraphics[width=1\linewidth]{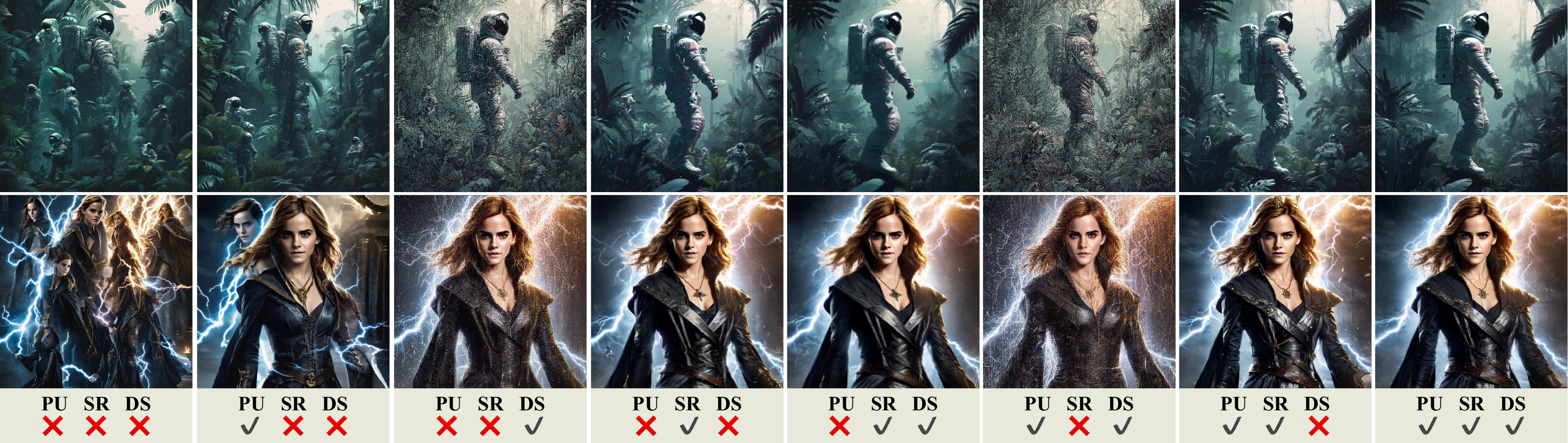}
\caption{\textbf{Ablation studies} on the three components of DemoFusion: Progressive Upscaling (PU), Skip Residual (SR), and Dilated Upsampling (DS). All images are generated at $3072^2$ ($9\times$ resolutions). Best viewed \textbf{ZOOMED-IN}.}
\label{fig:ablation}
\vspace{-0.3cm}
\end{figure*}

\section{Experiments}

Here, we report qualitative and quantitative experiments and ablation studies. For more details and results, please refer to Appendix: implementation details in Appendix~\ref{sec:implementation_details}, more discussions in Appendix~\ref{sec:discussion}, more visualisations in Appendix~\ref{sec:more_visualization}, more applications in Appendix~\ref{sec:more_application}, and all prompts we use in Appendix~\ref{sec:prompts}.

\subsection{Comparison}~\label{sec:comparison}
We compared DemoFusion with the following methods (i) \textbf{SDXL}~\cite{podell2023sdxl}, which is designed to generate images of $1024^2$. In the quantitative experiments, we also report the results of inferencing it at higher resolutions. (ii) \textbf{MultiDiffusion}~\cite{bar2023multidiffusion}, our baseline method based on overlapped local patch denoising. (iii) \textbf{SDXL+BSRGAN}. Using a super-resolution model is an intuitive solution to directly upscale SDXL results. Here, we choose BSRGAN~\cite{zhang2021designing}, a representative SR method, for comparison. (iv) \textbf{SCALECRAFTER}~\cite{he2023scalecrafter}, a concurrent training-free high-resolution generation method built on SDXL, which upscales by dilating convolutional kernels at specific layers. 

\noindent\textbf{Qualitative Results}: As shown in Fig.~\ref{fig:comparison}, each model is asked to generate images at $4\times$ and $16\times$ resolutions (compared to SDXL). We chose three prompts about realistic content rather than showcasing DemoFusion's prowess in artistic creation, as such content is more objective and facilitates a fair comparison.

Firstly, as previously mentioned, MultiDiffusion tends to generate repetitive content lacking semantic coherence. For SDXL+BSRGAN, we observe that the SR model effectively eliminates the blurriness and jagged edges of up-sampling, resulting in sharp and pleasing outcomes. However, the goal of the SR model is to produce images consistent with the input, which limits its performance in high-resolution generation -- needing more detail for true high-resolution visuals beyond simple smoothing. Checking the zoomed-in results of $4096^2$ -- compared to SDXL+BSRGAN, DemoFusion generates much richer details in the fur of the teddy bear, gives much richer details to Hermoine's eyes, and adds much more detail to the forest vegetation. This comparison confirms that high-resolution generation cannot be substituted by simple image super-resolution. As for SCALECRAFTER, while it partially addresses the issue of MultiDiffusion's repetitive content, it still needs improvement in semantic coherence. \emph{E.g.}, the teddy bear has multiple arms, eyes, or mouths. Additionally, directly dilating the convolutional kernels has somewhat affected the performance of the LDM, resulting in an overall image quality degradation, and local details exhibit many repetitive patterns (\emph{e.g.}, the trunks of the trees). In summary, the proposed DemoFusion achieves both rich local detail and strong global semantic coherence by modifying MultiDiffusion style patch-wise denoising paths to maximise the global context available for each path. 

\noindent\textbf{Quantitative Results}: For quantitative comparison, we adopt $3$ widely-used metrics: FID (Fréchet Inception Distance)~\cite{heusel2017gans}, IS (Inception Score)~\cite{salimans2016improved}, and CLIP Score~\cite{radford2021learning}. Considering that FID and IS require resizing images to $299^2$, which is not very suitable for high-resolution image assessment, inspired by~\cite{chai2022any}, we additionally crop local patches of $1\times$ resolution and then resize them to calculate these metrics, termed FID$_{crop}$ and IS$_{crop}$. The CLIP Score assesses the entire image's semantics; thus, we do not consider evaluating local patches here. We evaluate on the LAION-$5$B dataset~\cite{schuhmann2022laion} with $1K$ randomly sampled captions. Note that the results of FID and IS are related to the number of samples; therefore, the scores of FID$_{crop}$ and IS$_{crop}$ might be better than FID and IS due to more samples. The inference time is evaluated on an RTX $3090$ GPU.

As shown in Tab.~\ref{tbl:comparison}, DemoFusion achieved the best overall performance -- securing first or second place across all metrics. As the resolution increases, DemoFusion may score slightly lower than SDXL+BSRGAN on FID and IS because BSRGAN is designed to adhere strictly to low-resolution inputs, and these metrics also downsample images to low resolution for evaluation. However, DemoFusion significantly outperforms SDXL+BSRGAN on FID$_{crop}$ and IS$_{crop}$, indicating that DemoFusion can provide high-resolution local details. Besides, we observed that MultiDiffusion surpassed SCALECRAFTER on crop-based metrics due to these metrics' lack of an assessment of the overall structure of the image. Herefore, we keep the general FID and IS metrics. Regarding efficiency, since DemoFusion is based on MultiDiffusion and operates progressively, it requires a longer inference time. We discuss this point further in Sec.~\ref{sec:limitations}.

\begin{figure}[t]
\centering
\includegraphics[width=1\linewidth]{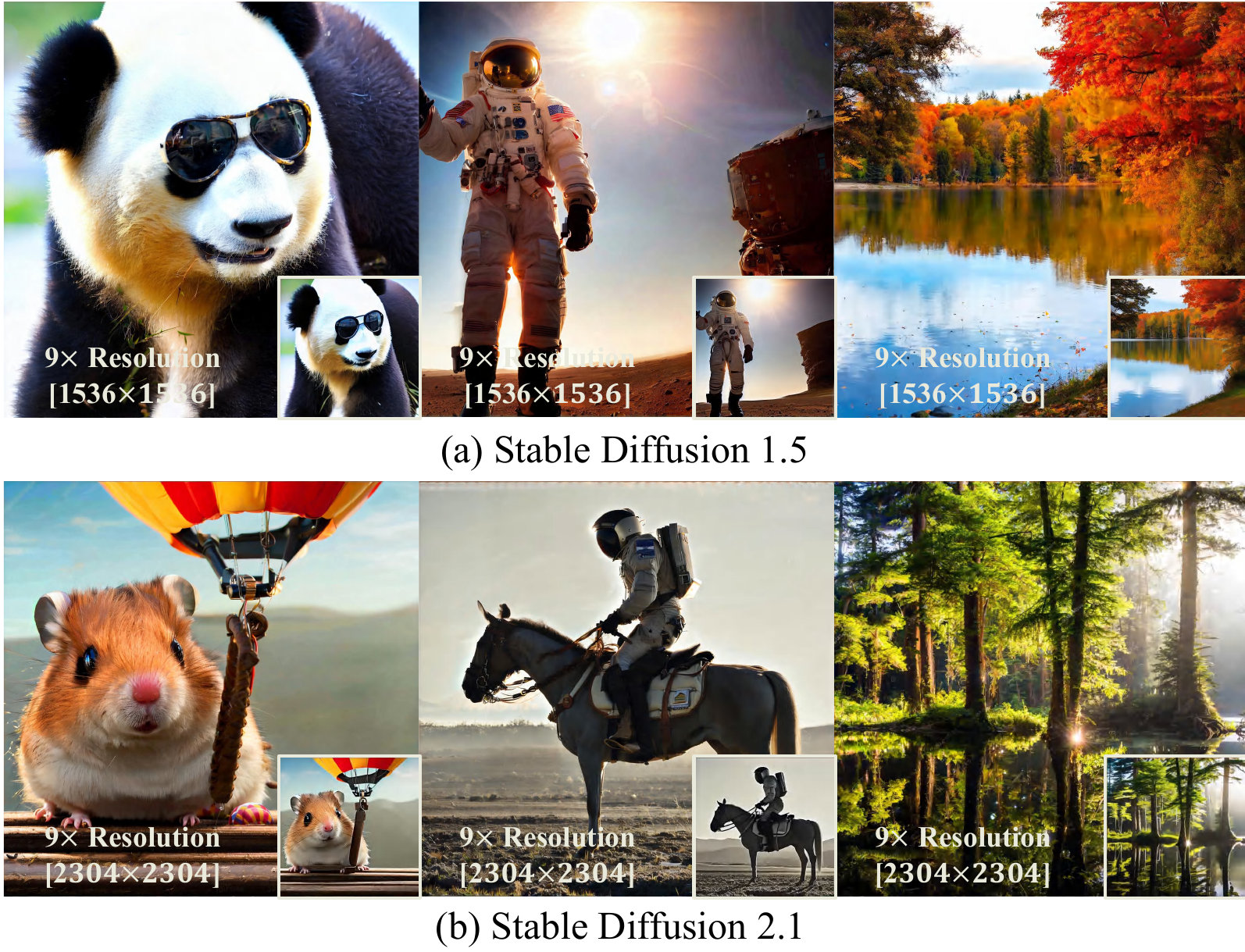}
\vspace{-0.6cm}
\caption{\textbf{Results of DemoFusion on other LDMs}, \emph{i.e.}, Stable Diffusion $1.5$ (default resolution of $512^2$) and Stable Diffusion $2.1$ (default resolution of $768^2$). All images are generated at $9\times$ resolutions. Best viewed \textbf{ZOOMED-IN}.}
\label{fig:other_ldm}
\vspace{-0.2cm}
\end{figure}

\begin{figure}[t]
\centering
\includegraphics[width=1\linewidth]{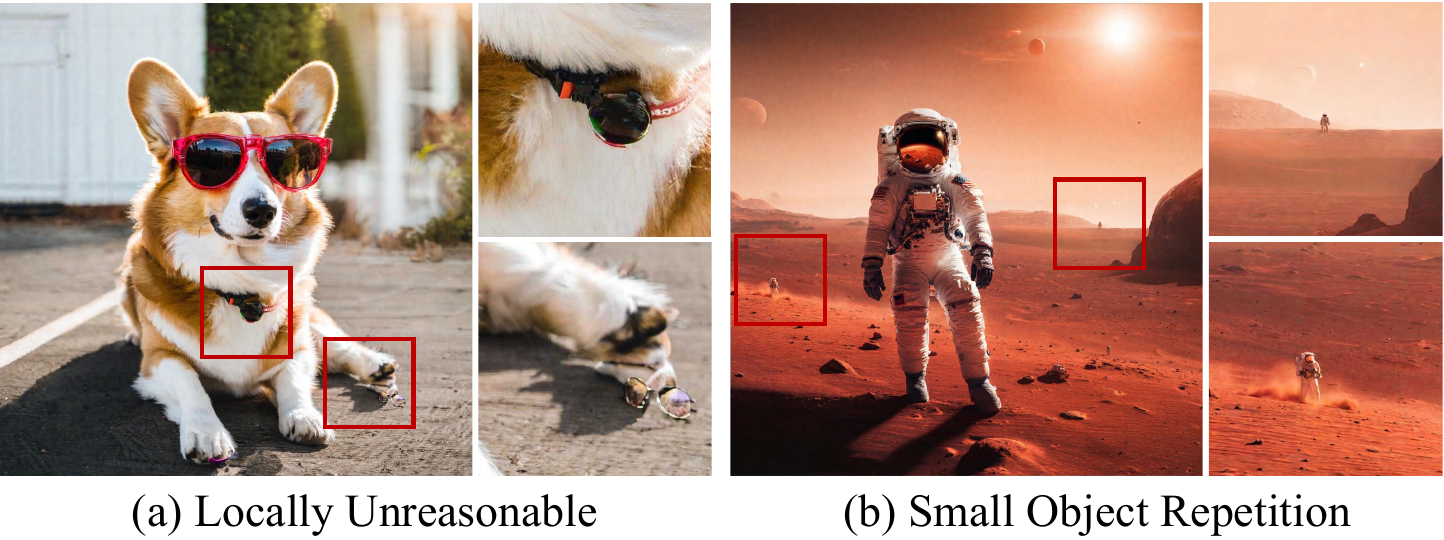}
\vspace{-0.6cm}
\caption{\textbf{Failure cases of DemoFusion}. (a) Irrational content appears locally in images with a sharp focus. (b) Small objects are repetitively present against a sparse background. All images are generated at $9\times$ resolutions. Best viewed \textbf{ZOOMED-IN}.}
\label{fig:failure_case}
\vspace{-0.6cm}
\end{figure}

\subsection{Ablation Study}
The proposed DemoFusion consists of three components: (i) progressive upscaling, (ii) skip residual, and (iii) dilated sampling. To visually demonstrate the effectiveness of these three components, we conducted experiments on all possible combinations, as shown in Fig.~\ref{fig:ablation}. All images are generated at $3072^2$ ($9\times$ resolutions). When all three components are removed, we generate at the original resolution first and then achieve higher resolutions via an ``upsample-diffuse-denoise'' loop. The results obtained under this setting are similar to naively generating via MultiDiffusion, with much repetitive content. However, this issue is gradually mitigated by incorporating the three proposed techniques, resulting in high-resolution images consistent with their original resolution counterparts.

Specifically, we found that continuously introducing information from the low resolution via skip residual dramatically helps maintain the overall structure to obtain acceptable results. On this basis, dilated sampling can further introduce denoising paths with global perception during the denoising process, guiding local denoising paths towards the global optimal direction. However, these mutually independent global denoising paths introduce two drawbacks (even though we have introduced Gaussian filtering to alleviate this): (i) bringing grainy textures when generating from Gaussian noises and (ii) amplifying the artificial noises introduced during the upscaling process. The former can be alleviated by introducing skip residuals, while the latter can be addressed by progressive upscaling, which prevents the strong artificial noises brought by direct large-scale upscaling. Overall, the three proposed techniques are complementary and indispensable. It is fascinating to see how well they work together.
\section{Limitations and Opportunities}~\label{sec:limitations}
DemoFusion exhibits limitations in the following aspects: (i) The nature of MultiDiffusion-style inference requires high computational load due to the overlapped denoising paths, and the progressive upscaling also prolongs inference times. (ii) As a tuning-free framework, DemoFusion's performance is directly correlated with the underlying LDM. In Fig.~\ref{fig:other_ldm}, we show the results based on other LDMs (Stable Diffusion $1.5$ and Stable Diffusion $2.1$), where DemoFusion is still effective, but the results are less astonishing than those on SDXL. (iii) DemoFusion entirely depends on the LDMs' prior knowledge of cropped images, and therefore, local irrational content may appear when generating sharp close-up images, as depicted in Fig.~\ref{fig:failure_case} (a). (iv) Although we have significantly mitigated the issue of repetitive content, the possibility of small repetitive content in background regions remains (see Fig.~\ref{fig:failure_case} (b)).

Behind these limitations, opportunities exist: (i) DemoFusion functions by fusing multiple denoising paths of the original size. This allows it to implement each denoising step in mini-batches, preventing the expected exponential increase in memory requirements. (ii) Although progressive upscaling requires more passes, users can acquire low-resolution intermediate results as ``previews'' within several seconds, facilitating rapid prompt iteration. (iii) The priors of current LDMs regarding image crops are solely derived from the general training scheme, which has already resulted in impressive performance. Training a bespoke LDM for a DemoFusion-like framework may be a promising direction to explore. 
\section{Conclusion}
In this paper, we introduce DemoFusion, a tuning-free framework that integrates plug-and-play with open-source GenAI models to achieve higher-resolution image generation. DemoFusion is built upon MultiDiffusion and introduces \emph{Progressive Upscaling}, \emph{Skip Residual}, and \emph{Dilated Sampling} techniques to enable generation with both global semantic coherence and rich local details.  DemoFusion persuasively demonstrates the possibility of LDMs generating images at higher resolutions than those used for training and the untapped potential of existing open-source GenAI models. By advancing the frontier of high-resolution image generation without additional training or prohibitive memory requirements for inference, we hope that DemoFusion can help democratize high-resolution image generation. 

%Our experiments demonstrate the outstanding performance of DemoFusion. We also discuss existing limitations and future opportunities. The most significant contribution of DemoFusion is its validation of the feasibility to train LDM models at a lower resolution for high-resolution generation. We hope that this work can propel the democratization process of high-resolution generation.
{
    \small
    \bibliographystyle{ieeenat_fullname}
    \bibliography{main}

\begin{thebibliography}{41}
\providecommand{\natexlab}[1]{#1}
\providecommand{\url}[1]{\texttt{#1}}
\expandafter\ifx\csname urlstyle\endcsname\relax
  \providecommand{\doi}[1]{doi: #1}\else
  \providecommand{\doi}{doi: \begingroup \urlstyle{rm}\Url}\fi

\bibitem[AI(2022)]{StableDiffusion2022}
Stability AI.
\newblock Stable diffusion: A latent text-to-image diffusion model.
\newblock \url{https://stability.ai/blog/stable-diffusion-public-release}, 2022.

\bibitem[Bar-Tal et~al.(2023)Bar-Tal, Yariv, Lipman, and Dekel]{bar2023multidiffusion}
Omer Bar-Tal, Lior Yariv, Yaron Lipman, and Tali Dekel.
\newblock Multidiffusion: Fusing diffusion paths for controlled image generation.
\newblock In \emph{ICML}, 2023.

\bibitem[Brooks et~al.(2023)Brooks, Holynski, and Efros]{brooks2023instructpix2pix}
Tim Brooks, Aleksander Holynski, and Alexei~A Efros.
\newblock Instructpix2pix: Learning to follow image editing instructions.
\newblock In \emph{CVPR}, 2023.

\bibitem[Chai et~al.(2022)Chai, Gharbi, Shechtman, Isola, and Zhang]{chai2022any}
Lucy Chai, Michael Gharbi, Eli Shechtman, Phillip Isola, and Richard Zhang.
\newblock Any-resolution training for high-resolution image synthesis.
\newblock In \emph{ECCV}, 2022.

\bibitem[Dhariwal and Nichol(2021)]{dhariwal2021diffusion}
Prafulla Dhariwal and Alexander Nichol.
\newblock Diffusion models beat gans on image synthesis.
\newblock In \emph{NeurIPS}, 2021.

\bibitem[Han et~al.(2023)Han, Cao, Han, Zhu, Deng, Song, Xiang, and Wong]{han2023headsculpt}
Xiao Han, Yukang Cao, Kai Han, Xiatian Zhu, Jiankang Deng, Yi-Zhe Song, Tao Xiang, and Kwan-Yee~K Wong.
\newblock Headsculpt: Crafting 3d head avatars with text.
\newblock In \emph{NeurIPS}, 2023.

\bibitem[He et~al.(2023{\natexlab{a}})He, Yang, Chen, Cun, Xia, Zhang, Wang, He, Chen, and Shan]{he2023scalecrafter}
Yingqing He, Shaoshu Yang, Haoxin Chen, Xiaodong Cun, Menghan Xia, Yong Zhang, Xintao Wang, Ran He, Qifeng Chen, and Ying Shan.
\newblock Scalecrafter: Tuning-free higher-resolution visual generation with diffusion models.
\newblock \emph{arXiv preprint arXiv:2310.07702}, 2023{\natexlab{a}}.

\bibitem[He et~al.(2023{\natexlab{b}})He, Yang, Zhang, Shan, and Chen]{he2023latent}
Yingqing He, Tianyu Yang, Yong Zhang, Ying Shan, and Qifeng Chen.
\newblock Latent video diffusion models for high-fidelity long video generation.
\newblock \emph{arXiv preprint arXiv:2211.13221}, 2023{\natexlab{b}}.

\bibitem[Hertz et~al.(2022)Hertz, Mokady, Tenenbaum, Aberman, Pritch, and Cohen-or]{hertz2022prompt}
Amir Hertz, Ron Mokady, Jay Tenenbaum, Kfir Aberman, Yael Pritch, and Daniel Cohen-or.
\newblock Prompt-to-prompt image editing with cross-attention control.
\newblock In \emph{ICLR}, 2022.

\bibitem[Heusel et~al.(2017)Heusel, Ramsauer, Unterthiner, Nessler, and Hochreiter]{heusel2017gans}
Martin Heusel, Hubert Ramsauer, Thomas Unterthiner, Bernhard Nessler, and Sepp Hochreiter.
\newblock Gans trained by a two time-scale update rule converge to a local nash equilibrium.
\newblock \emph{NeurIPS}, 2017.

\bibitem[Ho et~al.(2022{\natexlab{a}})Ho, Chan, Saharia, Whang, Gao, Gritsenko, Kingma, Poole, Norouzi, Fleet, et~al.]{ho2022imagen}
Jonathan Ho, William Chan, Chitwan Saharia, Jay Whang, Ruiqi Gao, Alexey Gritsenko, Diederik~P Kingma, Ben Poole, Mohammad Norouzi, David~J Fleet, et~al.
\newblock Imagen video: High definition video generation with diffusion models.
\newblock \emph{arXiv preprint arXiv:2210.02303}, 2022{\natexlab{a}}.

\bibitem[Ho et~al.(2022{\natexlab{b}})Ho, Saharia, Chan, Fleet, Norouzi, and Salimans]{ho2022cascaded}
Jonathan Ho, Chitwan Saharia, William Chan, David~J Fleet, Mohammad Norouzi, and Tim Salimans.
\newblock Cascaded diffusion models for high fidelity image generation.
\newblock \emph{The Journal of Machine Learning Research}, 2022{\natexlab{b}}.

\bibitem[Hoogeboom et~al.(2023)Hoogeboom, Heek, and Salimans]{hoogeboom2023simple}
Emiel Hoogeboom, Jonathan Heek, and Tim Salimans.
\newblock simple diffusion: End-to-end diffusion for high resolution images.
\newblock \emph{arXiv preprint arXiv:2301.11093}, 2023.

\bibitem[Jain et~al.(2023)Jain, Xie, and Abbeel]{jain2023vectorfusion}
Ajay Jain, Amber Xie, and Pieter Abbeel.
\newblock Vectorfusion: Text-to-svg by abstracting pixel-based diffusion models.
\newblock In \emph{CVPR}, 2023.

\bibitem[Karras et~al.(2018)Karras, Aila, Laine, and Lehtinen]{karras2018progressive}
Tero Karras, Timo Aila, Samuli Laine, and Jaakko Lehtinen.
\newblock Progressive growing of gans for improved quality, stability, and variation.
\newblock In \emph{ICLR}, 2018.

\bibitem[Lee et~al.(2023)Lee, Kim, Kim, and Sung]{lee2023syncdiffusion}
Yuseung Lee, Kunho Kim, Hyunjin Kim, and Minhyuk Sung.
\newblock Syncdiffusion: Coherent montage via synchronized joint diffusions.
\newblock \emph{arXiv preprint arXiv:2306.05178}, 2023.

\bibitem[Liao et~al.(2023)Liao, Yi, Xiu, Tang, Huang, Thies, and Black]{liao2023tada}
Tingting Liao, Hongwei Yi, Yuliang Xiu, Jiaxaing Tang, Yangyi Huang, Justus Thies, and Michael~J Black.
\newblock Tada! text to animatable digital avatars.
\newblock \emph{arXiv preprint arXiv:2308.10899}, 2023.

\bibitem[Lin et~al.(2023)Lin, Gao, Tang, Takikawa, Zeng, Huang, Kreis, Fidler, Liu, and Lin]{lin2023magic3d}
Chen-Hsuan Lin, Jun Gao, Luming Tang, Towaki Takikawa, Xiaohui Zeng, Xun Huang, Karsten Kreis, Sanja Fidler, Ming-Yu Liu, and Tsung-Yi Lin.
\newblock Magic3d: High-resolution text-to-3d content creation.
\newblock In \emph{CVPR}, 2023.

\bibitem[MidJourney(2022)]{MidJourney2022}
MidJourney.
\newblock Midjourney: An independent research lab.
\newblock \url{https://www.midjourney.com/}, 2022.

\bibitem[Mokady et~al.(2023)Mokady, Hertz, Aberman, Pritch, and Cohen-Or]{mokady2023null}
Ron Mokady, Amir Hertz, Kfir Aberman, Yael Pritch, and Daniel Cohen-Or.
\newblock Null-text inversion for editing real images using guided diffusion models.
\newblock In \emph{CVPR}, 2023.

\bibitem[Mou et~al.(2023)Mou, Wang, Xie, Zhang, Qi, Shan, and Qie]{mou2023t2i}
Chong Mou, Xintao Wang, Liangbin Xie, Jian Zhang, Zhongang Qi, Ying Shan, and Xiaohu Qie.
\newblock T2i-adapter: Learning adapters to dig out more controllable ability for text-to-image diffusion models.
\newblock \emph{arXiv preprint arXiv:2302.08453}, 2023.

\bibitem[OpenAI(2021)]{OpenAI2021DALL-E}
OpenAI.
\newblock Dall·e: Creating images from text.
\newblock \url{https://openai.com/blog/dall-e/}, 2021.

\bibitem[OpenAI(2022)]{OpenAI2021ChatGPT}
OpenAI.
\newblock Chatgpt: Large-scale language models.
\newblock \url{https://www.openai.com/blog/chatgpt}, 2022.

\bibitem[Podell et~al.(2023)Podell, English, Lacey, Blattmann, Dockhorn, M{\"u}ller, Penna, and Rombach]{podell2023sdxl}
Dustin Podell, Zion English, Kyle Lacey, Andreas Blattmann, Tim Dockhorn, Jonas M{\"u}ller, Joe Penna, and Robin Rombach.
\newblock Sdxl: Improving latent diffusion models for high-resolution image synthesis.
\newblock \emph{arXiv preprint arXiv:2307.01952}, 2023.

\bibitem[Poole et~al.(2022)Poole, Jain, Barron, and Mildenhall]{poole2022dreamfusion}
Ben Poole, Ajay Jain, Jonathan~T Barron, and Ben Mildenhall.
\newblock Dreamfusion: Text-to-3d using 2d diffusion.
\newblock In \emph{ICLR}, 2022.

\bibitem[Qu et~al.(2023)Qu, Xiang, and Song]{qu2023sketchdreamer}
Zhiyu Qu, Tao Xiang, and Yi-Zhe Song.
\newblock Sketchdreamer: Interactive text-augmented creative sketch ideation.
\newblock In \emph{BMVC}, 2023.

\bibitem[Radford et~al.(2021)Radford, Kim, Hallacy, Ramesh, Goh, Agarwal, Sastry, Askell, Mishkin, Clark, et~al.]{radford2021learning}
Alec Radford, Jong~Wook Kim, Chris Hallacy, Aditya Ramesh, Gabriel Goh, Sandhini Agarwal, Girish Sastry, Amanda Askell, Pamela Mishkin, Jack Clark, et~al.
\newblock Learning transferable visual models from natural language supervision.
\newblock In \emph{ICML}, 2021.

\bibitem[Rombach et~al.(2022)Rombach, Blattmann, Lorenz, Esser, and Ommer]{rombach2022high}
Robin Rombach, Andreas Blattmann, Dominik Lorenz, Patrick Esser, and Bj{\"o}rn Ommer.
\newblock High-resolution image synthesis with latent diffusion models.
\newblock In \emph{CVPR}, 2022.

\bibitem[Ruiz et~al.(2023)Ruiz, Li, Jampani, Pritch, Rubinstein, and Aberman]{ruiz2023dreambooth}
Nataniel Ruiz, Yuanzhen Li, Varun Jampani, Yael Pritch, Michael Rubinstein, and Kfir Aberman.
\newblock Dreambooth: Fine tuning text-to-image diffusion models for subject-driven generation.
\newblock In \emph{CVPR}, 2023.

\bibitem[Salimans et~al.(2016)Salimans, Goodfellow, Zaremba, Cheung, Radford, and Chen]{salimans2016improved}
Tim Salimans, Ian Goodfellow, Wojciech Zaremba, Vicki Cheung, Alec Radford, and Xi Chen.
\newblock Improved techniques for training gans.
\newblock \emph{NeurIPS}, 2016.

\bibitem[Schuhmann et~al.(2022)Schuhmann, Beaumont, Vencu, Gordon, Wightman, Cherti, Coombes, Katta, Mullis, Wortsman, et~al.]{schuhmann2022laion}
Christoph Schuhmann, Romain Beaumont, Richard Vencu, Cade Gordon, Ross Wightman, Mehdi Cherti, Theo Coombes, Aarush Katta, Clayton Mullis, Mitchell Wortsman, et~al.
\newblock Laion-5b: An open large-scale dataset for training next generation image-text models.
\newblock \emph{NeurIPS}, 2022.

\bibitem[Sohl-Dickstein et~al.(2015)Sohl-Dickstein, Weiss, Maheswaranathan, and Ganguli]{sohl2015deep}
Jascha Sohl-Dickstein, Eric Weiss, Niru Maheswaranathan, and Surya Ganguli.
\newblock Deep unsupervised learning using nonequilibrium thermodynamics.
\newblock In \emph{ICML}, 2015.

\bibitem[Song et~al.(2021)Song, Meng, and Ermon]{song2020denoising}
Jiaming Song, Chenlin Meng, and Stefano Ermon.
\newblock Denoising diffusion implicit models.
\newblock In \emph{ICLR}, 2021.

\bibitem[Wang et~al.(2023{\natexlab{a}})Wang, Yue, Zhou, Chan, and Loy]{wang2023exploiting}
Jianyi Wang, Zongsheng Yue, Shangchen Zhou, Kelvin~CK Chan, and Chen~Change Loy.
\newblock Exploiting diffusion prior for real-world image super-resolution.
\newblock \emph{arXiv preprint arXiv:2305.07015}, 2023{\natexlab{a}}.

\bibitem[Wang et~al.(2023{\natexlab{b}})Wang, Zhang, Zhang, Gu, Bao, Baltrusaitis, Shen, Chen, Wen, Chen, et~al.]{wang2023rodin}
Tengfei Wang, Bo Zhang, Ting Zhang, Shuyang Gu, Jianmin Bao, Tadas Baltrusaitis, Jingjing Shen, Dong Chen, Fang Wen, Qifeng Chen, et~al.
\newblock Rodin: A generative model for sculpting 3d digital avatars using diffusion.
\newblock In \emph{CVPR}, 2023{\natexlab{b}}.

\bibitem[Wu et~al.(2023)Wu, Ge, Wang, Lei, Gu, Shi, Hsu, Shan, Qie, and Shou]{wu2023tune}
Jay~Zhangjie Wu, Yixiao Ge, Xintao Wang, Stan~Weixian Lei, Yuchao Gu, Yufei Shi, Wynne Hsu, Ying Shan, Xiaohu Qie, and Mike~Zheng Shou.
\newblock Tune-a-video: One-shot tuning of image diffusion models for text-to-video generation.
\newblock In \emph{CVPR}, 2023.

\bibitem[Xu et~al.(2023)Xu, Wang, Cheng, Cao, Shan, Qie, and Gao]{xu2023dream3d}
Jiale Xu, Xintao Wang, Weihao Cheng, Yan-Pei Cao, Ying Shan, Xiaohu Qie, and Shenghua Gao.
\newblock Dream3d: Zero-shot text-to-3d synthesis using 3d shape prior and text-to-image diffusion models.
\newblock In \emph{CVPR}, 2023.

\bibitem[Yu and Vladlen(2016)]{yu2018multi}
Fisher Yu and Koltun Vladlen.
\newblock Multi-scale context aggregation by dilated convolutions.
\newblock In \emph{ICLR}, 2016.

\bibitem[Zhang et~al.(2021)Zhang, Liang, Van~Gool, and Timofte]{zhang2021designing}
Kai Zhang, Jingyun Liang, Luc Van~Gool, and Radu Timofte.
\newblock Designing a practical degradation model for deep blind image super-resolution.
\newblock In \emph{CVPR}, 2021.

\bibitem[Zhang et~al.(2023)Zhang, Rao, and Agrawala]{zhang2023adding}
Lvmin Zhang, Anyi Rao, and Maneesh Agrawala.
\newblock Adding conditional control to text-to-image diffusion models.
\newblock In \emph{ICCV}, 2023.

\bibitem[Zheng et~al.(2023)Zheng, Guo, Deng, Han, Li, Xu, and Xu]{zheng2023any}
Qingping Zheng, Yuanfan Guo, Jiankang Deng, Jianhua Han, Ying Li, Songcen Xu, and Hang Xu.
\newblock Any-size-diffusion: Toward efficient text-driven synthesis for any-size hd images.
\newblock \emph{arXiv preprint arXiv:2308.16582}, 2023.

\end{thebibliography}
}

% WARNING: do not forget to delete the supplementary pages from your submission 
\cleardoublepage
\begin{appendix}
\renewcommand{\thesection}{\Alph{section}}
\section*{Appendix}

\begin{algorithm*}[htbp]
\caption{Image Synthesis Process of DemoFusion}
\begin{algorithmic}[1]
\State \texttt{\#\#\#\#\#\#\#\#\#\#\#\#\#\#\#\#\#\#\#\#\#\#\#\#\#\# Phase $1$ \#\#\#\#\#\#\#\#\#\#\#\#\#\#\#\#\#\#\#\#\#\#\#\#\#\#\#}
\State $\mathbf{z}^0_T \sim \mathcal{N}(0, I)$ \Comment{Random Initialization}
\For{$t = T$ to $1$}
    \State $p_\theta(\mathbf{z}^1_{t-i}|\mathbf{z}^1_t)$ \Comment{Denoising Step}
\EndFor
\State \texttt{\#\#\#\#\#\#\#\#\#\#\#\#\#\#\#\#\#\#\#\#\#\#\#\# Phase $2$ to $S$ \#\#\#\#\#\#\#\#\#\#\#\#\#\#\#\#\#\#\#\#\#\#\#\#}
\For{$s = 2$ to $S$}
    \State $inter(\mathbf{z'}^s_0|\mathbf{z}^{s-1}_0)$ \Comment{Upsampling}
    \For{$t = 1$ to $T$}
        \State $q(\mathbf{z'}^s_t|\mathbf{z'}^s_{t-1})$ \Comment{Diffusion Step}
    \EndFor
    \For{$t = T$ to $1$}
        \State $\mathbf{\hat{z}}^s_t = c_1 \times \mathbf{z'}^s_t + (1-c_1) \times \mathbf{z}^s_t$ \Comment{Skip Residual}
        \State $\mathcal{S}_{local}(\mathbf{\hat{z}}^s_t) \rightarrow Z_{t}^{local}$ \Comment{Crop Sampling (MultiDiffusion)}
        \State $\mathcal{S}_{global}(\mathbf{\hat{z}}^s_t) \rightarrow Z_{t}^{global}$ \Comment{Dilated Sampling}
        \For{$\mathbf{\hat{z}}^s_{n,t}$ in $Z_{t}^{local}$}
            \State $p_\theta(\mathbf{z}^s_{n,t-i}|\mathbf{\hat{z}}^s_{n,t})$ \Comment{Local Path Denoising Step (MultiDiffusion)}
        \EndFor
        \For{$\mathbf{\hat{z}}^s_{m,t}$ in $Z_{t}^{global}$}
            \State $p_\theta(\mathbf{z}^s_{m,t-i}|\mathbf{\hat{z}}^s_{m,t})$ \Comment{Global Path Denoising Step}
        \EndFor
        \State $\mathcal{R}_{local}(Z_{t-1}^{local}) \times (1 - c_2) + \mathcal{R}_{global}(Z_{t-1}^{global}) \times c_2 \rightarrow \mathbf{z}^s_t$ \Comment{Fusing Local and Global Paths}
    \EndFor
\EndFor
\State \textbf{return} $\mathbf{x}^S_0=\mathcal{D}(\mathbf{z}^S_0)$ 
\Comment{Decoding to Image}
\end{algorithmic}
\label{alg:demofusion}
\end{algorithm*}

\begin{figure}[htbp]
\centering
\includegraphics[width=1\linewidth]{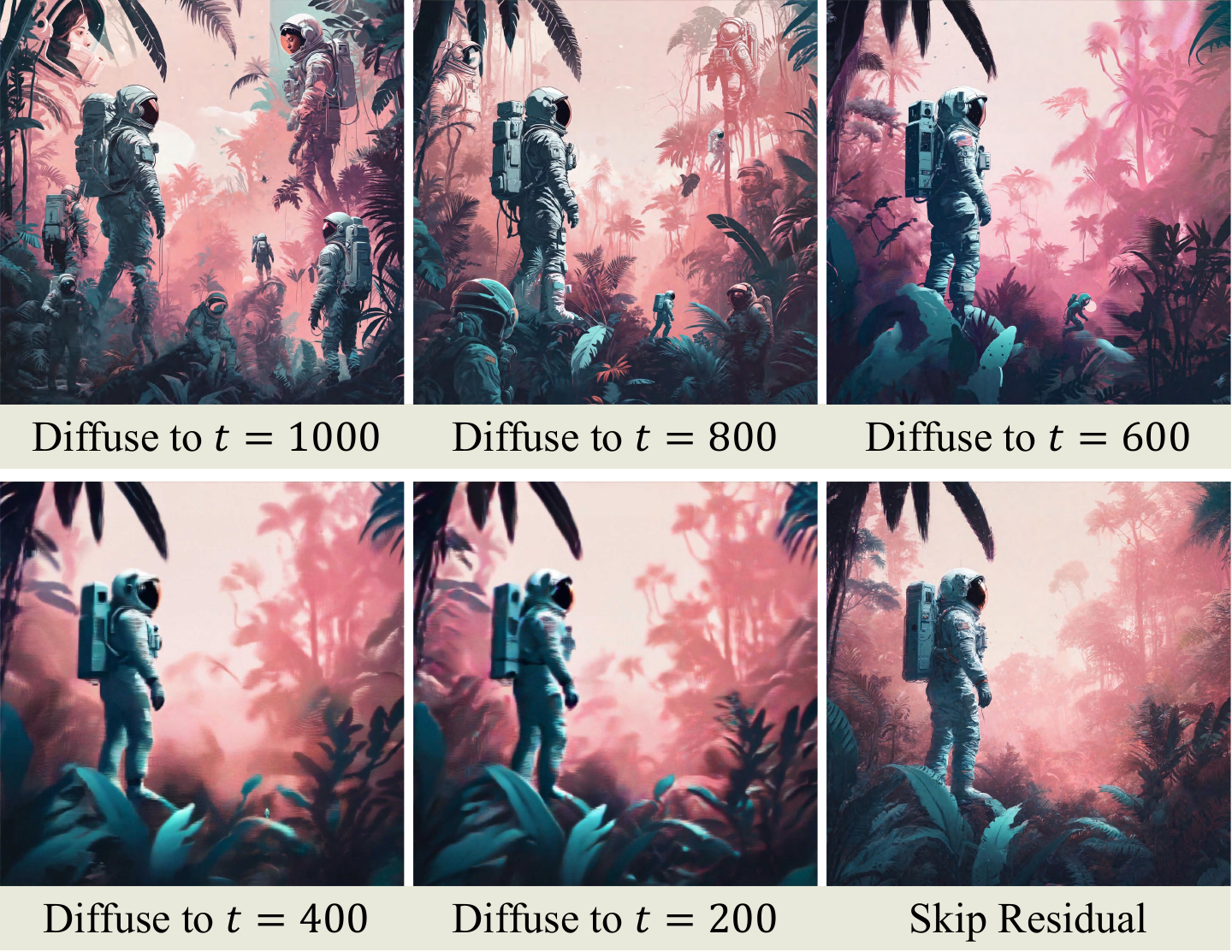}
\caption{
During the progressive upscaling process, we diffuse $\textbf{z}^s_0$ to different time-steps $t$, and then denoise it back to obtain the results. The number of training time-step is $1000$.}
\label{fig:different_t}
\vspace{-0.4cm}
\end{figure}

\begin{table}[htpb]
% 	\vspace{-1.0cm} 
\renewcommand{\arraystretch}{1.3}
\centering
\begin{adjustbox}{width=1\linewidth,center}
\begin{tabular}{ccc|ccccc}
\toprule[1pt]
PU & SR & DS & FID & IS & FID$_{crop}$ & IS$_{crop}$ & CLIP \\
% \midrule[0.5pt]
% \multicolumn{19}{c}{\emph{\textbf{Greedy-Hungarian}}}\\
\midrule[1pt]
$\mathbf{\times}$ & $\mathbf{\times}$ & $\mathbf{\times}$ & $95.28$ & $13.92$ & $80.11$ & $19.61$ & $28.13$ \\
$\mathbf{\checkmark}$ & $\mathbf{\times}$ & $\mathbf{\times}$ & $90.77$ & $13.95$ & $79.23$ & $20.08$ & $28.52$ \\
$\mathbf{\times}$ & $\mathbf{\checkmark}$ & $\mathbf{\times}$ & $79.93$ & $15.19$ & $74.17$ & $22.48$ & $29.40$ \\
$\mathbf{\times}$ & $\mathbf{\times}$ & $\mathbf{\checkmark}$ & $94.35$ & $14.89$ & $82.32$ & $19.64$ & $28.85$ \\
$\mathbf{\checkmark}$ & $\mathbf{\checkmark}$ & $\mathbf{\times}$ & $75.92$ & $15.66$ & $72.98$ & $23.20$ & $29.50$ \\
$\mathbf{\checkmark}$ & $\mathbf{\times}$ & $\mathbf{\checkmark}$ & $89.26$ & $15.02$ & $80.04$ & $21.86$ & $28.87$ \\
$\mathbf{\times}$ & $\mathbf{\checkmark}$ & $\mathbf{\checkmark}$ & $76.53$ & $15.71$ & $73.22$ & $23.09$ & $29.48$ \\
$\mathbf{\checkmark}$ & $\mathbf{\checkmark}$ & $\mathbf{\checkmark}$ & $\mathbf{74.11}$ & $\mathbf{16.11}$ & $\mathbf{70.34}$ & $\mathbf{24.28}$ & $\mathbf{29.57}$ \\
\midrule[1pt]
\end{tabular}
\end{adjustbox}
\caption{\textbf{Quantitative results of the ablation study.} The best results are marked in \textbf{bold}. Impact of components: Progressive Upscaling (PU), Skip Residual (SR), and Dilated
Upsampling (DS). }
\label{tbl:ablation}
\vspace{-0.4cm}
\end{table}

\begin{figure}[h]
\centering
\includegraphics[width=1\linewidth]{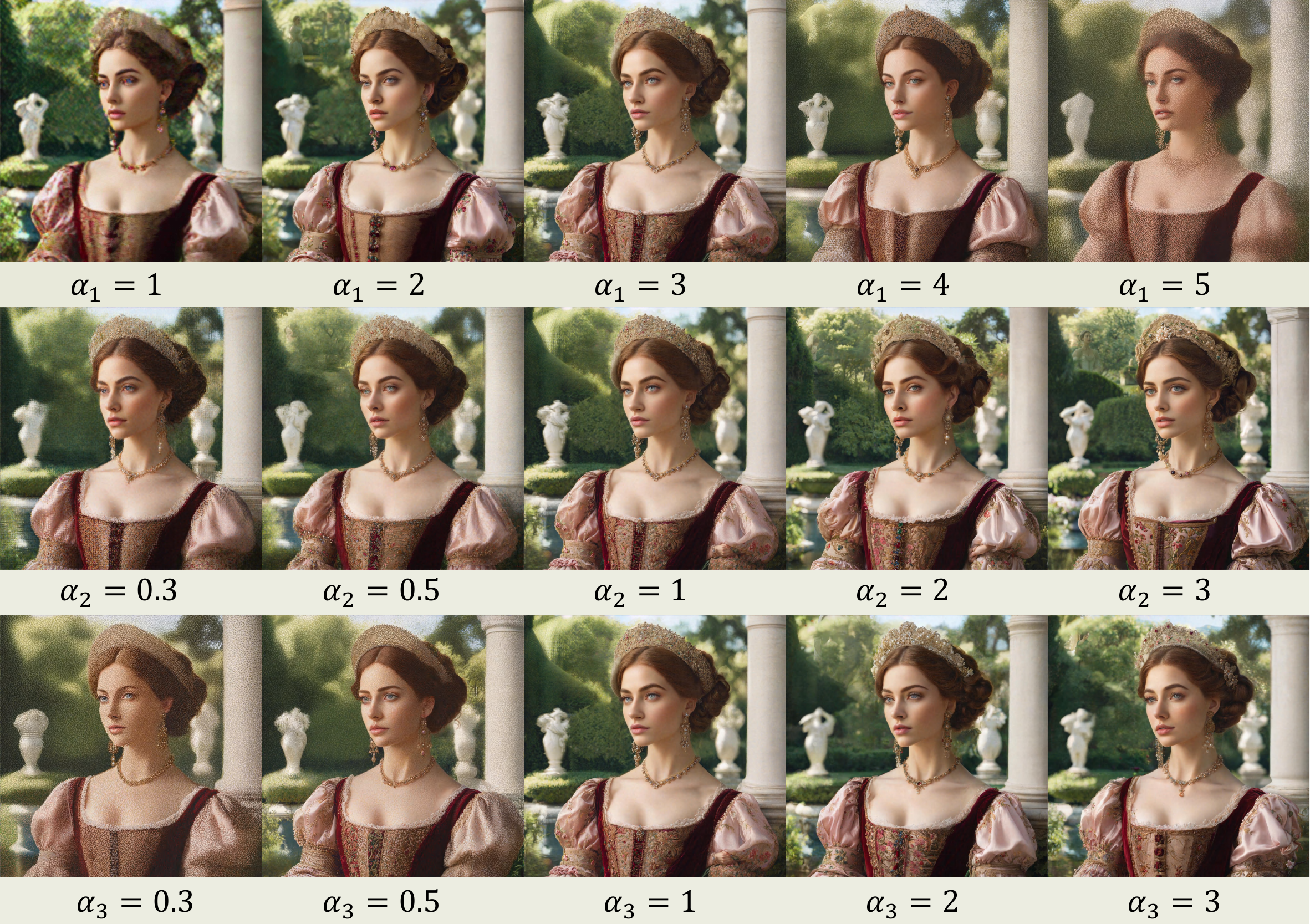}
\caption{\textbf{Results with different $\alpha_1$, $\alpha_2$, and $\alpha_3$.} All images are generated at $3072^2$ ($9\times$ resolutions). Best viewed \textbf{ZOOMED-IN}.}
\label{fig:hyper-parameters}
\vspace{-0.4cm}
\end{figure}

\begin{figure*}[htpb]
\centering
\includegraphics[width=1\linewidth]{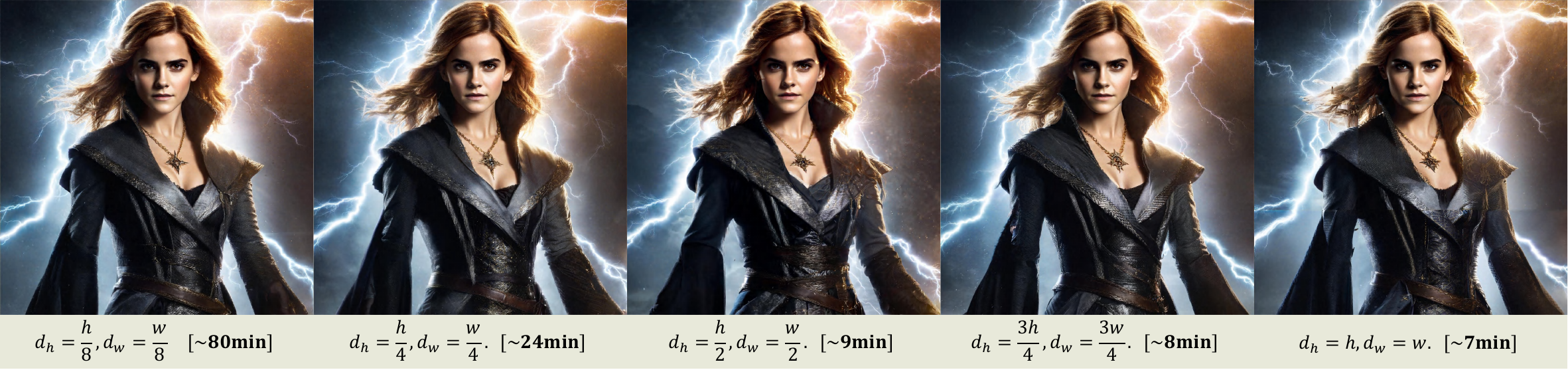}
\caption{\textbf{Results with different strides $d_h$ and $d_w$}. All images are generated at $3072^2$ ($9\times$ resolutions). Best viewed \textbf{ZOOMED-IN}.}
\label{fig:different_d}
\vspace{-0.4cm}
\end{figure*}

\begin{figure}[htpb]
\centering
\includegraphics[width=1\linewidth]{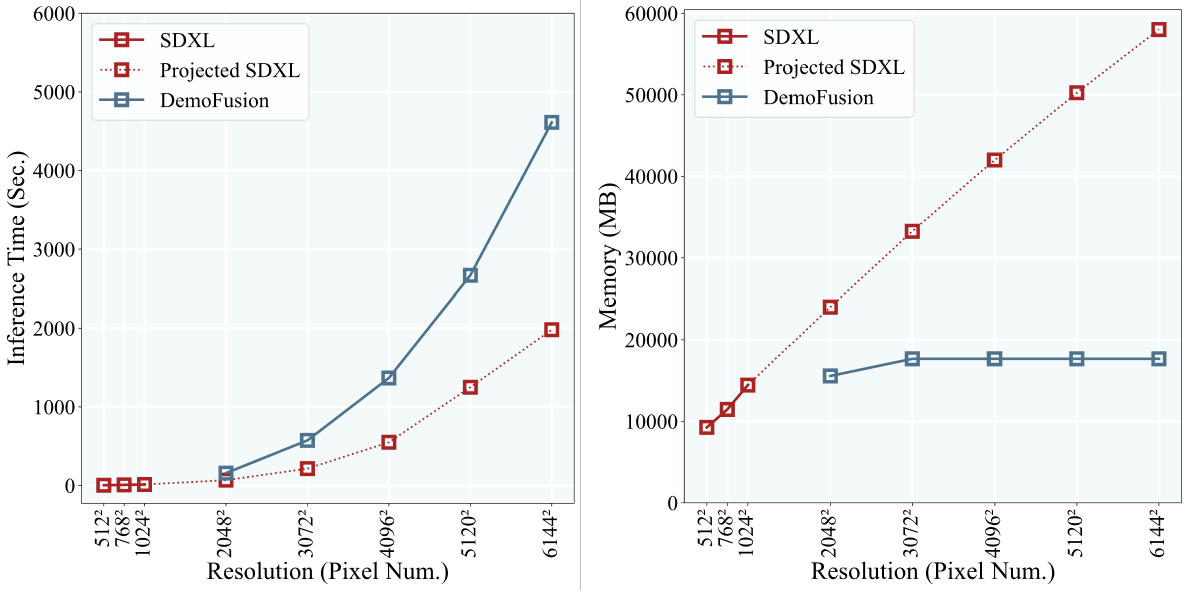}
\caption{Inference time of SDXL \emph{versus} DemoFusion (\textbf{Left}). Memory demands of SDXL \emph{versus} DemoFusion (\textbf{Right}).}
\label{fig:time_memory}
\vspace{-0.4cm}
\end{figure}

\begin{figure*}[htpb]
\centering
\includegraphics[width=1\linewidth]{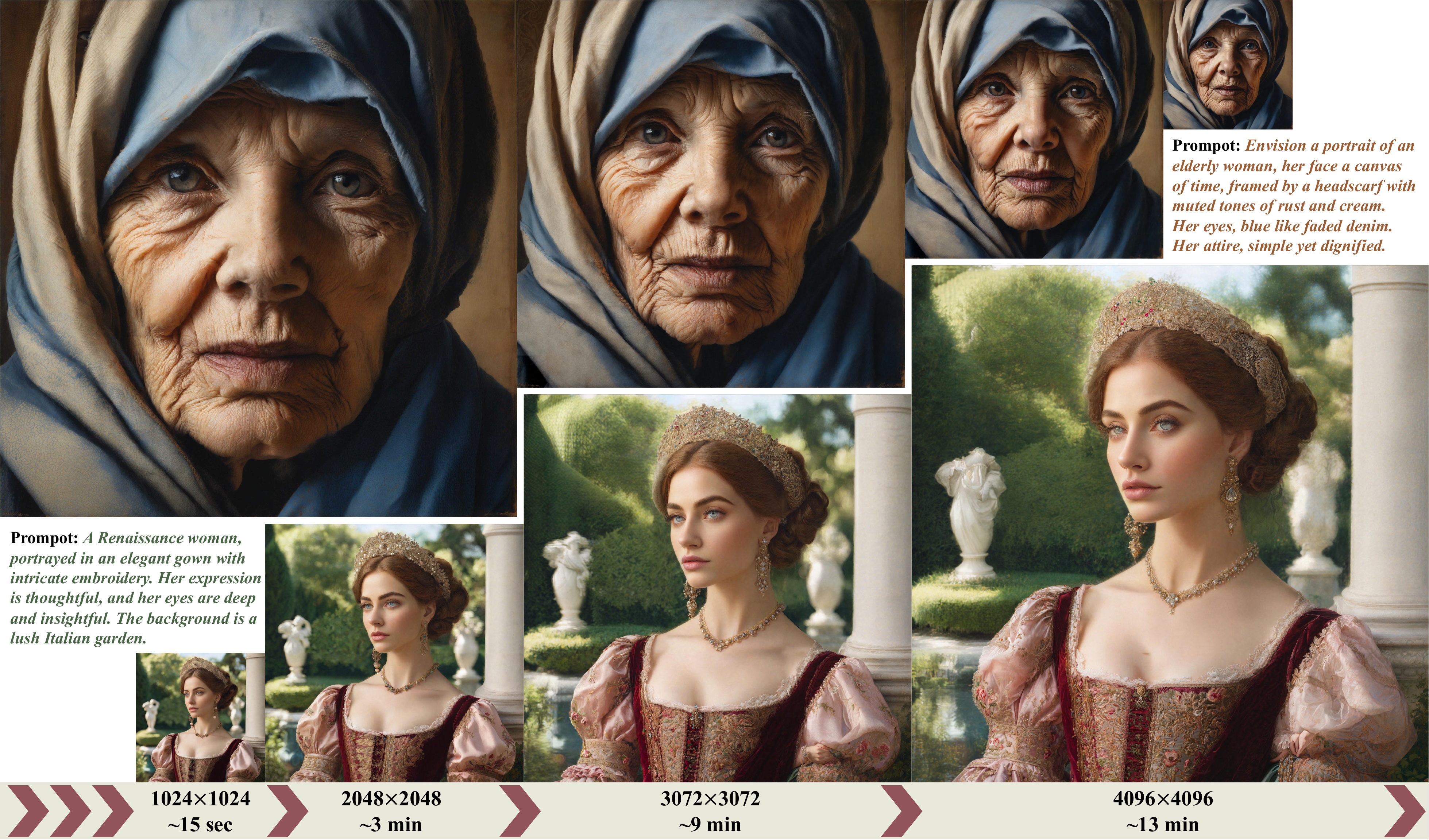}
\caption{\textbf{Illustration of the progressive upscaling process.} The time required for each phase is indicated. Best viewed \textbf{ZOOMED-IN}.}
\label{fig:progressive_process}
\vspace{-0.2cm}
\end{figure*}

\begin{figure*}[htbp]
\centering
\includegraphics[width=1\linewidth]{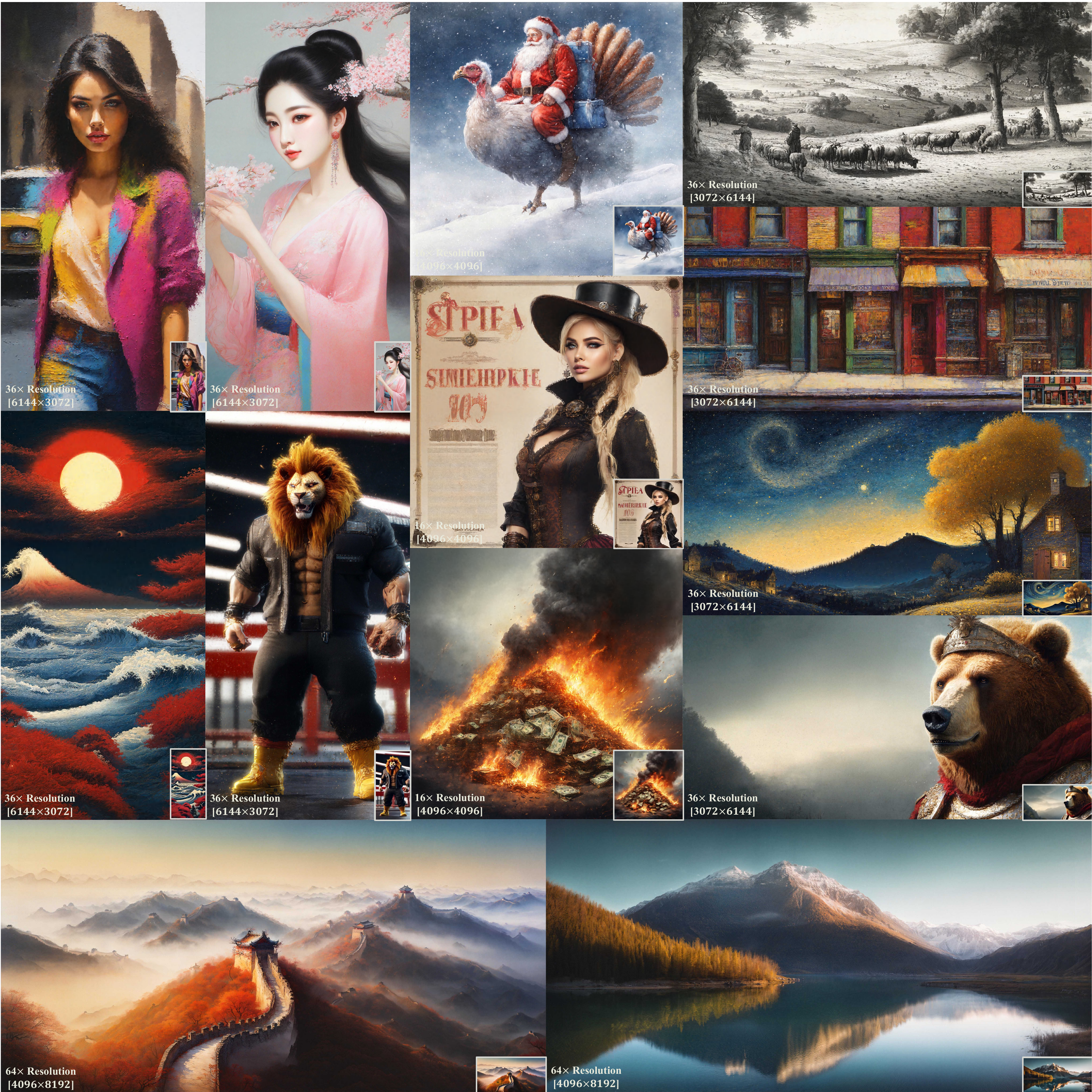}
\caption{\textbf{More selected landscape samples of DemoFusion \emph{versus} SDXL}~\cite{podell2023sdxl} (all images in the figure are presented at their actual sizes). All generated images are produced using a single RTX $3090$ GPU. Best viewed \textbf{ZOOMED-IN}.}
\label{fig:illustration2}
\vspace{-0.4cm}
\end{figure*}

\begin{figure*}[htbp]
\centering
\includegraphics[width=1\linewidth]{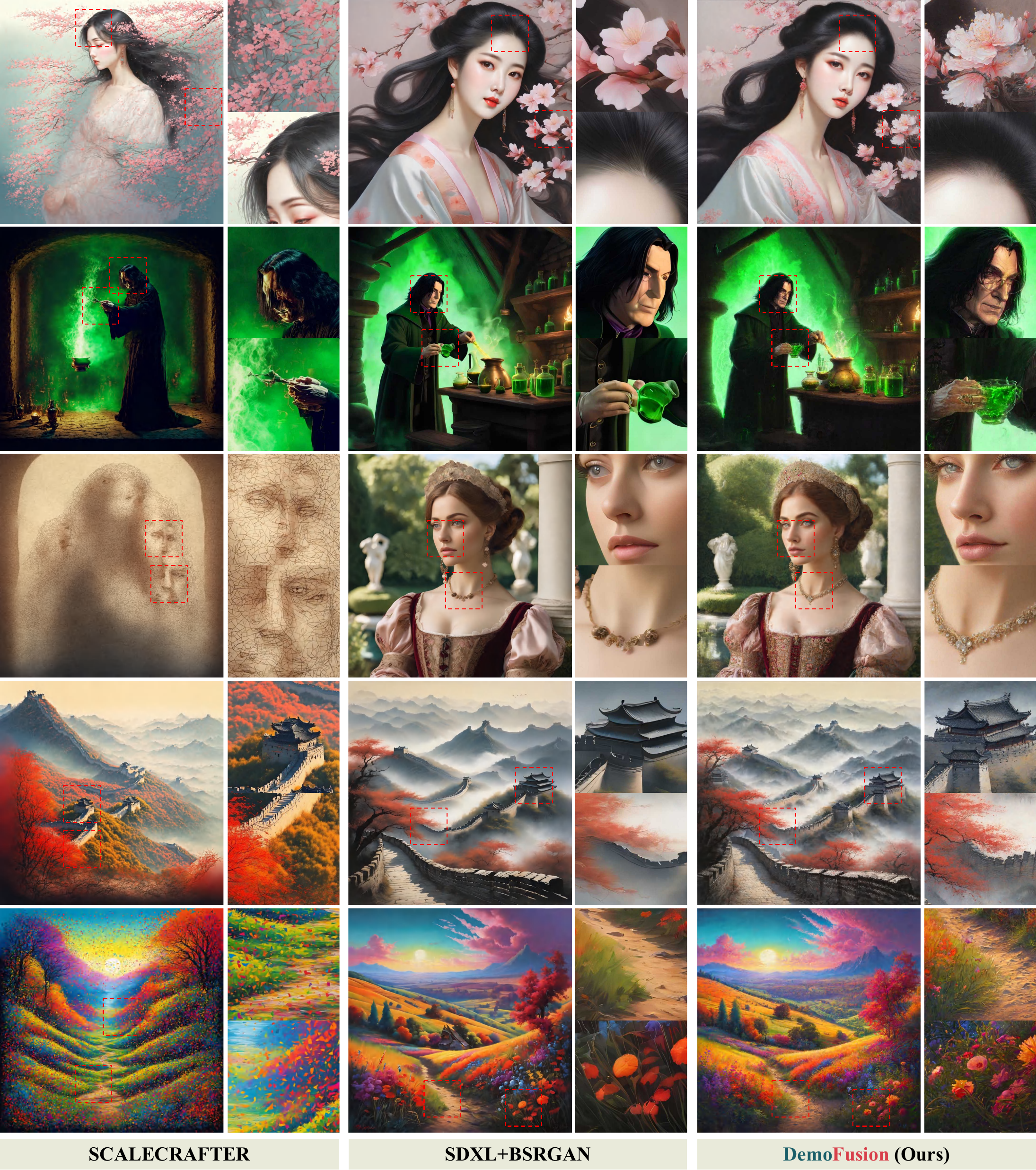}
\caption{\textbf{More Qualitative comparison results.} All images are generated at $4096^2$ ($16\times$ resolutions). Local details have already been zoomed in, but it’s still recommended to \textbf{ZOOM IN} for a closer look.}
\label{fig:comparison2}
\vspace{0.4cm}
\end{figure*}

\begin{figure*}[htbp]
\centering
\includegraphics[width=1\linewidth]{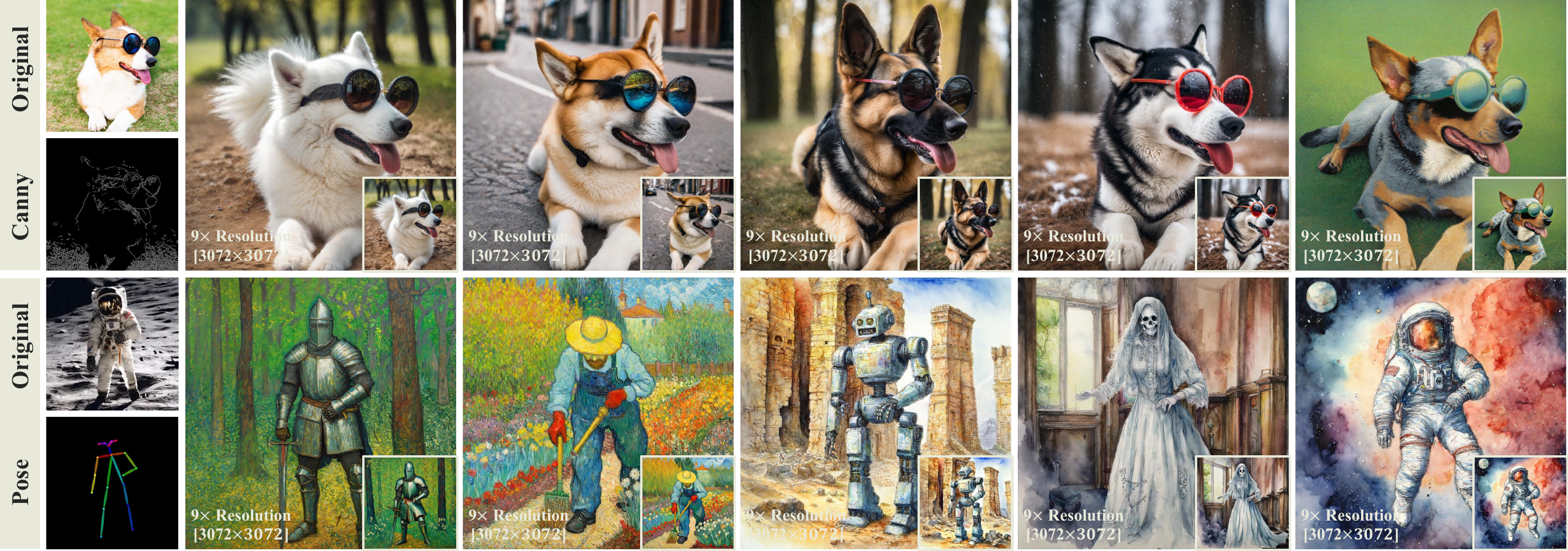}
\caption{\textbf{Results of DemoFusion combining with ControlNet~\cite{zhang2023adding}}. All images are generated at $3072^2$ ($9\times$ resolutions). Best viewed \textbf{ZOOMED-IN}.}
\label{fig:controlnet}
\vspace{-0.4cm}
\end{figure*}

\begin{figure}[htbp]
\centering
\includegraphics[width=1\linewidth]{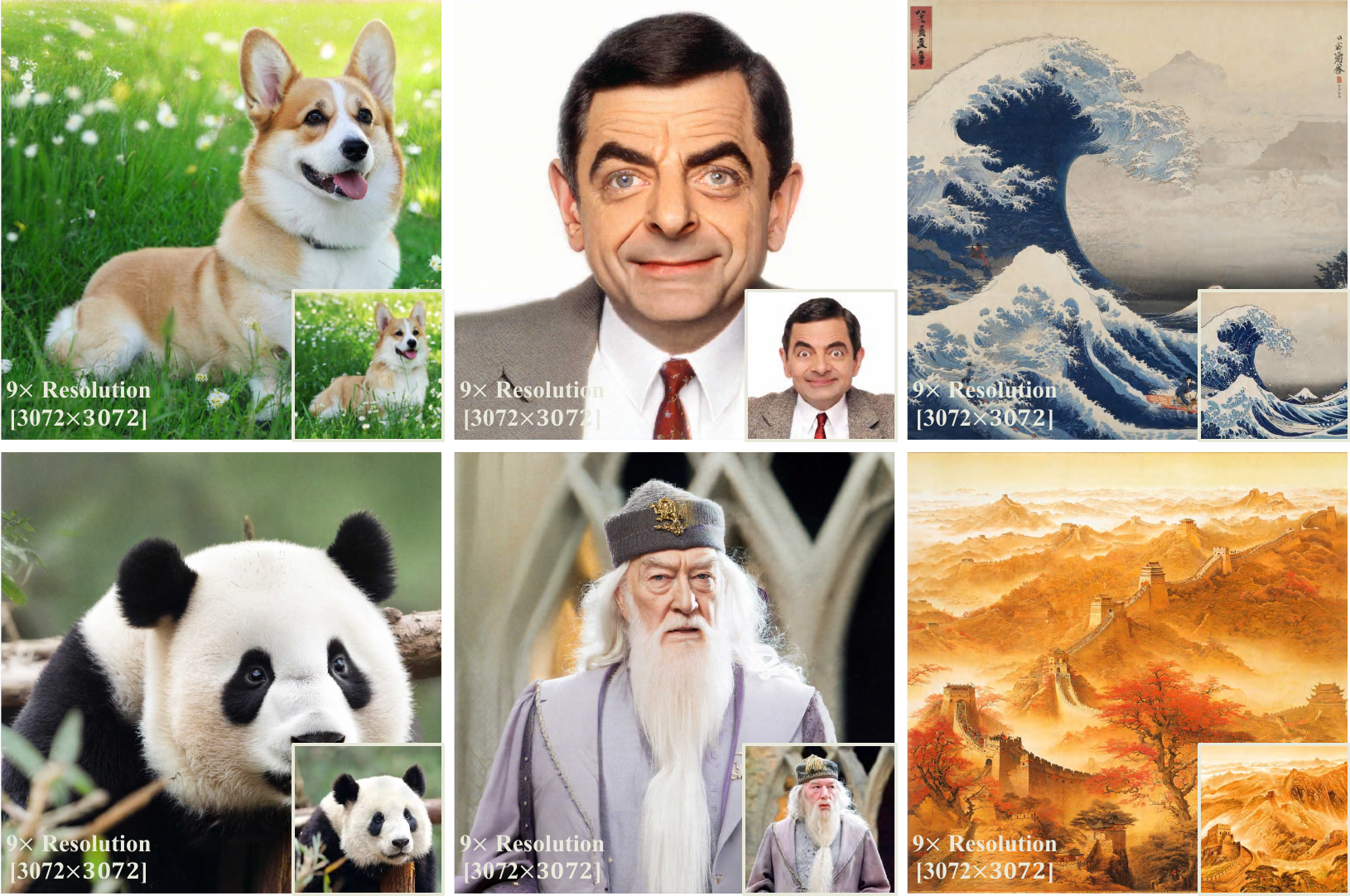}
\caption{\textbf{Results of upscaling real images}. All images are upsacled to $3072^2$. Best viewed \textbf{ZOOMED-IN}.}
\label{fig:real_iamge}
\vspace{-0.4cm}
\end{figure}

\section{Pseudo Code}~\label{sec:pseudo_code}
We further illustrate the image synthesis process of DemoFusion in Algorithm~\ref{alg:demofusion}.

\section{Implementation Details}~\label{sec:implementation_details}
In cases where it is not explicitly stated, all the results in this paper are obtained based on SDXL with a DDIM scheduler of $50$ steps. The guidance scale for all denoising paths is set to $7.5$. The crop size of MultiDiffusion is set to be aligned with the maximum training size of pre-trained LDMs, \emph{e.g.}, $h=w=128$ for SDXL, and the stride is set to be $d_h=\frac{h}{2}$ and $d_w=\frac{w}{2}$. Each crop's position is subjected to a slight random perturbation, with maximum offsets of $\frac{h}{16}$ and $\frac{w}{16}$ in vertical and horizontal directions, respectively, further preventing the occurrence of seam issues.

When generating images with varying aspect ratios, we ensure that the longer side aligns with the maximum training size. Three scale factors $\alpha_1$, $\alpha_2$, $\alpha_3$ were set to $3$, $1$, and $1$ respectively. The Gaussian filter's standard deviation decreases from $\sigma_1=1$ to $\sigma_2=0.01$. To decode high-resolution images, we also employed a tiled decoder strategy as \cite{he2023scalecrafter} and some open-source projects\footnote{\url{https://github.com/pkuliyi2015/multidiffusion-upscaler-for-automatic1111}}. To eliminate seams between tiles, we sample a larger range of features around each tile during the decoding process.

Note that SDXL permits the input of a coarse cropping condition~\cite{podell2023sdxl}, \emph{i.e.}, the coordinates of the top-left corner of the cropping area. Therefore, when utilizing SDXL, we additionally input the coordinates of the top-left corner of the corresponding patch as a condition for local denoising paths. However, we observed that the presence or absence of this condition does not significantly impact the results. Besides, DemoFusion initiates the generation process from the highest resolution of the LDM during the first phase. When generating images with varying aspect ratios, we ensure that the longer side aligns with the highest resolution.

\section{More Experimental Results}~\label{sec:discussion}
\vspace{-0.5cm}
\subsection{Diffusing to Different Time-step $t$}
In Fig.~\ref{fig:different_t}, we illustrate that, when skip residual is removed, the effects of different time-steps we denoise to within the ``upsample-diffuse-denoise'' loop. This provides evidence for our discussion in the main text -- the larger the $t$, the more information is lost, which weakens the global perception; the smaller the $t$, the stronger the noise introduced by upsampling.

\subsection{Quantitative Results of Ablation Study}
The quantitative results of the ablation study are shown in Tab.~\ref{tbl:ablation}. Here, we only experiment with the resolution of $4096^2$.

\subsection{Effects of Scale Factors $\alpha_1$, $\alpha_2$, and $\alpha_3$}
A shared understanding of the DM's denoising process is that the DM first determines the coarse details and then gradually refines the local details. In line with this understanding, we adopt a unified strategy: utilizing cosine descending weights, we assign greater weights to skip residuals, dilated sampling, and accompanying Gaussian filtering in the early stages of the denoising process, gradually decreasing the weights as denoising progresses. Despite this unified approach, the three components still need distinct scale factors to control the descent rate.

Through grid search, we obtained the globally optimal parameter combination. In Fig.~\ref{fig:hyper-parameters}, we varied only one parameter at a time while keeping the others at their optimal values to demonstrate the impact of each parameter on the results. Note that a larger scale factor means a faster decline, which weakens the effect of this item, and vice versa. 

According to the experimental results, when the skip residual effect is too strong (\emph{i.e.}, $\alpha_1=1$), we observe significant artificial noise caused by upsampling. Because these factors interact with each other, when $\alpha_1=5$, we observe that the results are close to the one when $\alpha_3=1$ -- Gaussian filtering leads to excessive smoothing of latent representation. The trade-off of dilated sampling is – too large a weight (\emph{i.e.}, $\alpha_2=1$) can result in grainy images, while too small a weight (\emph{i.e.}, $\alpha_2=5$) fails to provide sufficient global perception, leading to noticeable issues of content repetition. Regarding Gaussian filtering, excessive strength can lead to over-smoothing of the latent representation, while too small a strength can weaken the global denoising paths due to lack of interaction, resulting in content repetition and grainy appearance.

\subsection{Effect of Stride Sizes $d_h$ and $d_w$}
In general, the stride size $d_h$ and $d_w$ in MultiDiffusion determines the extent of the seam issue of images -- a smaller stride means more seamless images, but at the same time, brings more overlapping computation. For DemoFusion, due to the proposed \emph{Progressive Upscaling}, \emph{Skip Residual}, and \emph{Dilated Sampling techniques}, there is a better consistency between patches, relaxing the stride size requirement. In Fig.~\ref{fig:different_d}, we showcase the performance of DemoFusion and the corresponding inference time for different stride sizes. It can be observed that even in the case of $d_h=h$ and $d_w=w$, \emph{i.e.}, no overlap between patches, we still achieve good global semantic coherence, while when $d_h<h$ and $d_w<w$, all the generated images have no noticeable seams; ultimately, in order to balance the performance and efficiency, we chose $d_h=\frac{h}{2}$ and $d_w=\frac{w}{2}$.

\subsection{Resource Demands of DemoFusion}
In Fig.~\ref{fig:time_memory}, we illustrate the resource demand comparison of DemoFusion and the original SDXL~\cite{podell2023sdxl}. Note that SDXL cannot generate valid content at resolutions higher than $1024^2$. We just calculate the expected resource demands by assuming we have a high-resolution SDXL under the current framework. It can be seen that DemoFusion achieves high-resolution image generation on limited computational resources while paying a little bit more time cost.

\section{More Visualizations}~\label{sec:more_visualization}
\vspace{-0.5cm}
\subsection{The Progressive Upscaling Process}
To better demonstrate how the model progressively generates images with different resolutions, in Fig.~\ref{fig:progressive_process}, we show the model outputs at each phase of the generation process. We can observe that DemoFusion does an excellent job of achieving global consistency under different resolutions, indicating the reason for its success in resolving content repetition.

\subsection{More Landscape Samples}
In Fig.~\ref{fig:illustration2}, we have supplemented more samples to show the performance of DemoFusion, and in particular, we further show results at the resolution of $8192\times4096$ ($64\times$ upscaling compared to the initial resolution of $512\times1024$).

\subsection{More Comparison Results}
In Fig.~\ref{fig:comparison2}, we have supplemented more comprison results with \textbf{SDXL+BSRGAN}~\cite{zhang2021designing} and \textbf{SCALECRAFTER}~\cite{he2023scalecrafter} to demonstrate the effectiveness of DemoFusion. The results are consistent with those in the main text. Compared to SDXL+BSRGAN, DemoFusion provides better local details; while compared to SCALECRAFTER, DemoFusion better preserves the performance of SDXL during upscale.

\section{More Applications}~\label{sec:more_application}
\vspace{-0.5cm}
\subsection{Combining with ControlNet}
The tuning-free characteristic of DemoFusion enables seamless integration with many LDM-based applications. \emph{E.g.}, DemoFusion combined with ControlNet~\cite{zhang2023adding} can achieve controllable high-resolution generation. In Fig.~\ref{fig:controlnet}, we showcase examples using Canny edge and human pose as conditions.

\subsection{Upscaling Real Images}
Since DemoFusion works in a progressive manner, we can replace the output of phase $1$ with representations obtained by encoding real images, thereby achieving upscaling of real images. However, we carefully avoid using the term ``super resolution'', as the outputs tend to lean towards the latent data distribution of the base LDM, making this process more akin to image generation based on a real image. The results are shown in the Fig.~\ref{fig:real_iamge}.

\section{Prompts Used in This Paper}~\label{sec:prompts}
All prompts used in this paper are taken from the internet or generated by ChatGPT~\cite{OpenAI2021ChatGPT}. They are summarised here.

\noindent\textbf{Fig.~\ref{fig:illustration} in the main text:}
\begin{itemize}
    \item \textit{Steampunk makeup, in the style of vray tracing, colorful impasto, uhd image, indonesian art, fine feather details with bright red and yellow and green and pink and orange colours, intricate patterns and details, dark cyan and amber makeup. Rich colourful plumes. Victorian style.}
    \item \textit{Stunning feminine body, commercial image, beautiful girl from Spain, holographic photography shoots, large body of water sprayed, liquid splashing all over the places, street pop, luminous palette, close up, realistic impressionism, shiny/glossy, extreme colorsplash, behind that a universe of vortex of fire waves and ice waves, around fire splashes and ice splashes and floral, bonsais, roots, smoke swirls, dust swirls, tentacles of fire and ice, s-curve composition, leading lines, cinematic, style of hokusai, unreal engine, octane render, asymetric, golden ratio, style of hokusai, liquid splashes, merging, melting, splashing, droplets, mixing, fading away, exploding, swirling, intricate detail, modelshoot style, dreamlikeart, dramatic lighting. 8k, highly detailed, trending artstation.}
    \item \textit{The beautiful scenery of Seattle, painting by Al Capp.}
    \item \textit{By Tang Yau Hoong, ultra hd, realistic, vivid colors, highly detailed, UHD drawing, pen and ink, perfect composition, beautiful detailed intricate insanely detailed octane render trending on artstation, 8k artistic photography, photorealistic concept art, soft natural volumetric cinematic perfect light, ultra hd, realistic, vivid colors, highly detailed, UHD drawing, pen and ink, perfect composition, beautiful detailed intricate insanely detailed octane render trending on artstation, 8k artistic photography, photorealistic concept art, soft natural volumetric cinematic perfect light.}
    \item \textit{A cute and adorable fluffy puppy wearing a witch hat in a halloween autumn evening forest, falling autumn leaves, brown acorns on the ground, halloween pumpkins spiderwebs, bats, a witch's broom.}
    \item \textit{A robot standing in the rain reading newspaper, rusty and worn down, in a dystopian cyberpunk street, photo-realistic, urbanpunk.}
    \item \textit{Einstein, a bronze statue, with a fresh red apple on his head, by Bruno Catalano.}
    \item \textit{A woman in a pink dress walking down a street, cyberpunk art, inspired by Victor Mosquera, conceptual art, style of raymond swanland, yume nikki, restrained, robot girl, ghost in the shell.}
    \item \textit{Photo of a rhino dressed suit and tie sitting at a table in a bar with a bar stools, award winning photography, Elke vogelsang.}
    \item \textit{An astronaut riding a horse on the moon, oil painting by Van Gogh.}
    \item \textit{Classic traditional cornucopia at the fall harvest festival, farm in the background, high quality masterful still-life painting, American pastoral, oil painting, festive spirit, vibrant cultural tradition, Autumnal atmosphere, vibrant rich colors.}
\end{itemize}

\noindent\textbf{Fig.~\ref{fig:intuition} in the main text:}
\begin{itemize}
    \item \textit{An astronaut riding a horse on the moon, oil painting by Van Gogh.}
\end{itemize}

\noindent\textbf{Fig.~\ref{fig:framework} in the main text:}
\begin{itemize}
    \item \textit{An astronaut riding a horse on the moon, oil painting by Van Gogh.}
\end{itemize}

\noindent\textbf{Fig.~\ref{fig:comparison} in the main text:}
\begin{itemize}
    \item \textit{A cute teddy bear in front of a plain white wall, warm and brown fur, soft and fluffy.}
    \item \textit{Emma Watson as a powerful mysterious sorceress, casting lightning magic, detailed clothing.}
    \item \textit{Primitive forest, towering trees, sunlight falling, vivid colors.}
\end{itemize}

\noindent\textbf{Fig.~\ref{fig:ablation} in the main text:}
\begin{itemize}
    \item \textit{Astronaut in a jungle, cold color palette, muted colors, detailed, 8k.}
    \item \textit{Emma Watson as a powerful mysterious sorceress, casting lightning magic, detailed clothing.}
\end{itemize}

\noindent\textbf{Fig.~\ref{fig:other_ldm} in the main text:}
\begin{itemize}
    \item \textit{A panda wearing sunglasses.}
    \item \textit{Astronaut on Mars During sunset.}
    \item \textit{A serene lakeside during autumn, with trees displaying a palette of fiery colors.}
    \item \textit{A hamster piloting a tiny hot air balloon.}
    \item \textit{An astronaut riding a horse.}
    \item \textit{A deep forest clearing with a mirrored pond reflecting a galaxy-filled night sky.}
\end{itemize}

\noindent\textbf{Fig.~\ref{fig:failure_case} in the main text:}
\begin{itemize}
    \item \textit{A corgi wearing cool sunglasses.}
    \item \textit{Astronaut on Mars During sunset.}
\end{itemize}

\noindent\textbf{Fig.~\ref{fig:different_t} in Appendix:}
\begin{itemize}
    \item \textit{Astronaut in a jungle, cold color palette, muted colors, detailed, 8k.}
\end{itemize}

\noindent\textbf{Fig.~\ref{fig:hyper-parameters} in Appendix:}
\begin{itemize}
    \item \textit{A Renaissance noblewoman, portrayed in an elegant gown with intricate embroidery. Her expression is thoughtful, and her eyes are deep and insightful. The background is a lush Italian garden, reflecting the artistic style of the High Renaissance.}
\end{itemize}

\noindent\textbf{Fig.~\ref{fig:different_d} in Appendix:}
\begin{itemize}
    \item \textit{Emma Watson as a powerful mysterious sorceress, casting lightning magic, detailed clothing.}
\end{itemize}

\noindent\textbf{Fig.~\ref{fig:progressive_process} in Appendix:}
\begin{itemize}
    \item \textit{Envision a portrait of an elderly woman, her face a canvas of time, framed by a headscarf with muted tones of rust and cream. Her eyes, blue like faded denim. Her attire, simple yet dignified.}
    \item \textit{A Renaissance noblewoman, portrayed in an elegant gown with intricate embroidery. Her expression is thoughtful, and her eyes are deep and insightful. The background is a lush Italian garden, reflecting the artistic style of the High Renaissance.}
\end{itemize}

\noindent\textbf{Fig.~\ref{fig:illustration2} in Appendix:}
\begin{itemize}
    \item \textit{Realistic oil painting of a stunning model merged in multicolor splash made of finely torn paper, eye contact, walking with class in a street.}
    \item \textit{A painting of a beautiful graceful woman with long hair, a fine art painting, by Qiu Ying, no gradients, flowing sakura silk, beautiful oil painting.}
    \item \textit{Katsushika Hokusai's Japanese depiction of a very turbulent sea with massive waves. The background s shows a beautiful dark night over a illuminated village. The colors are red and yellow, mood lighting Imagine a dreamlike scene blending the swirling cosmic colors of Vincent van Gogh's Starry Night with the surreal celestial precision of Salvador Dalí.}
    \item \textit{Character of lion in style of saiyan, mafia, gangsta, citylights background, Hyper detailed, hyper realistic, unreal engine ue5, cgi 3d, cinematic shot, 8k.}
    \item \textit{Santa Claus riding on top of a turkey, with very large bag of gifts, snow, ice, very cold place, realistic digital art, blurred background, expansive lighting, 4k, light gray and blue color palette, sharp and fine intricate details defined.}
    \item \textit{Best Quality, Masterpiece, steampunk theme, centered, front cover of fashion magazine, concept art, design, magazine design, 1girl, cute, blonde ponytail hair, gothic steampunk dress, model pose, (epic composition, epic proportion), vibrant color, text, diagrams, advertisements, magazine title, typography.}
    \item \textit{Burning pile of money, epic composition, digital painting, emotionally profound, thought-provoking, intense and brooding tones, high quality, masterpiece.}
    \item \textit{A pastoral scene with shepherds, flocks, and rolling hills, in the tradition of a Jean-François Millet landscape.}
    \item \textit{A painting of brooklyn new york 1940 storefronts, by John Kay, highly textured, rich colour and detail, ballard, deep colour\'s, style of raymond swanland, trio, oill painting, h 768, well worn, displayed, detailed 4 k oil painting, glenn barr, textured oil on canvas, looking cute.}
    \item \textit{A swirling night sky filled with bright stars and a small village below, inspired by Vincent van Gogh's Starry Night.}
    \item \textit{Portrait of a bear as a roman general, with a helmet, decorative, fantasy environment, oil painting, masterpiece, detailed, sharp, clear, cinematic lights.}
    \item \textit{The Great Wall of China winding through mist-covered mountains, captured in the delicate brushwork and harmonious colors of a traditional Chinese landscape painting.}
    \item \textit{RAW photo of a mountain lake landscape, clean water, 8k, UHD.}
\end{itemize}

\noindent\textbf{Fig.~\ref{fig:comparison2} in Appendix:}
\begin{itemize}
    \item \textit{A painting of a beautiful graceful woman with long hair, a fine art painting, by Qiu Ying, no gradients, flowing sakura silk, beautiful oil painting.}
    \item \textit{Professor Snape brewing a potion in the dungeon, the room illuminated by the green glow of the cauldron.}
    \item \textit{A Renaissance noblewoman, portrayed in an elegant gown with intricate embroidery. Her expression is thoughtful, and her eyes are deep and insightful. The background is a lush Italian garden, reflecting the artistic style of the High Renaissance.}
    \item \textit{The Great Wall of China winding through mist-covered mountains, captured in the delicate brushwork and harmonious colors of a traditional Chinese landscape painting.}
    \item \textit{Summer landscape, vivid colors, a work of art, grotesque, Mysterious.}
\end{itemize}

\noindent\textbf{Fig.~\ref{fig:controlnet} in Appendix:}
\begin{itemize}
    \item \textit{A Samoyed wearing a sunglasses, sticking out its tongue, dslr image, 8k.}
    \item \textit{A Corgi wearing a sunglasses, sticking out its tongue, dslr image, 8k.}
    \item \textit{A German Shepherd wearing a sunglasses, sticking out its tongue, on the grass, dslr image, 8k.}
    \item \textit{A Husky wearing a sunglasses, sticking out its tongue, dslr image, 8k.}
    \item \textit{An Australian Cattle Dog wearing a sunglasses, sticking out its tongue, style of Gian Lorenzo Bernini.}
    \item \textit{A medieval knight standing in a lush forest, oil painting by Van Gogh.}
    \item \textit{A gardener tending to a colorful, blooming garden, oil painting by Van Gogh.}
    \item \textit{A robot exploring the ruins of an ancient civilization, watercolor by MORILAND.}
    \item \textit{A ghost haunting an abandoned Victorian mansion, watercolor by MORILAND.}
    \item \textit{An astronaut floating in space, watercolor by MORILAND.}
\end{itemize}

\noindent\textbf{Fig.~\ref{fig:real_iamge} in Appendix:}
\begin{itemize}
    \item \textit{A cute corgi on the lawn.}
    \item \textit{A portrait of Mr. Bean (Rowan Atkinson).}
    \item \textit{Japanese Ukiyo-e, Kanagawa Surfing Sato.}
    \item \textit{A cute panda on a tree trunk.}
    \item \textit{A Portrait of Albus Dumbledore.}
    \item \textit{A Chinese Painting of the Great Wall.}
\end{itemize}
\end{appendix}

\end{document}